\documentclass[sigconf, nonacm]{acmart}
\usepackage{bm}
\usepackage{eso-pic} 
\usepackage{tabularx} 
\usepackage{booktabs} 
\usepackage{multirow}
\usepackage{ragged2e}
\usepackage{colortbl} % 用于表格颜色
\usepackage{enumitem} % 列表定制
\usepackage[most]{tcolorbox} %用于漂亮的彩色盒子
\definecolor{pinkbg}{HTML}{FFBBD0} % 亮粉色 (明亮而不刺眼)
\definecolor{bluebg}{HTML}{B1DDF1} % 亮蓝色 (清爽且识别度高)
\definecolor{mainPurple}{RGB}{155, 103, 168} 
\definecolor{SoftBlue}{RGB}{100, 149, 237} 
\definecolor{SoftPink}{RGB}{255, 182, 193} 
\definecolor{SoftPurple}{RGB}{180, 150, 210} 
\definecolor{SoftOrange}{RGB}{255, 210, 160} 
\newtcolorbox{custombox}[2]{
 colback=#1!3!white, 
 colframe=#1,  
 colbacktitle=#1,  
 coltitle=white,  
 fonttitle=\bfseries\large, 
 title={#2},  
 arc=3mm,   
 left=12pt, right=12pt, top=8pt, bottom=8pt,
 toptitle=3pt, bottomtitle=3pt,
 boxrule=1.5pt,  
}
\newcolumntype{L}{>{\RaggedRight\arraybackslash}X}
\newcolumntype{C}{>{\Centering\arraybackslash}X}
\begin{document}
% --- 新增：页面绝对定位代码放在这里 ---
\AddToShipoutPictureBG*{ 
 \AtPageUpperLeft{
 % 图片基准点在左下角：向下移 2.0cm，向右移 2cm (和页面边距对齐)
 \put(\LenToUnit{2cm}, \LenToUnit{-2.0cm}){ 
  \includegraphics[height=0.8 cm]{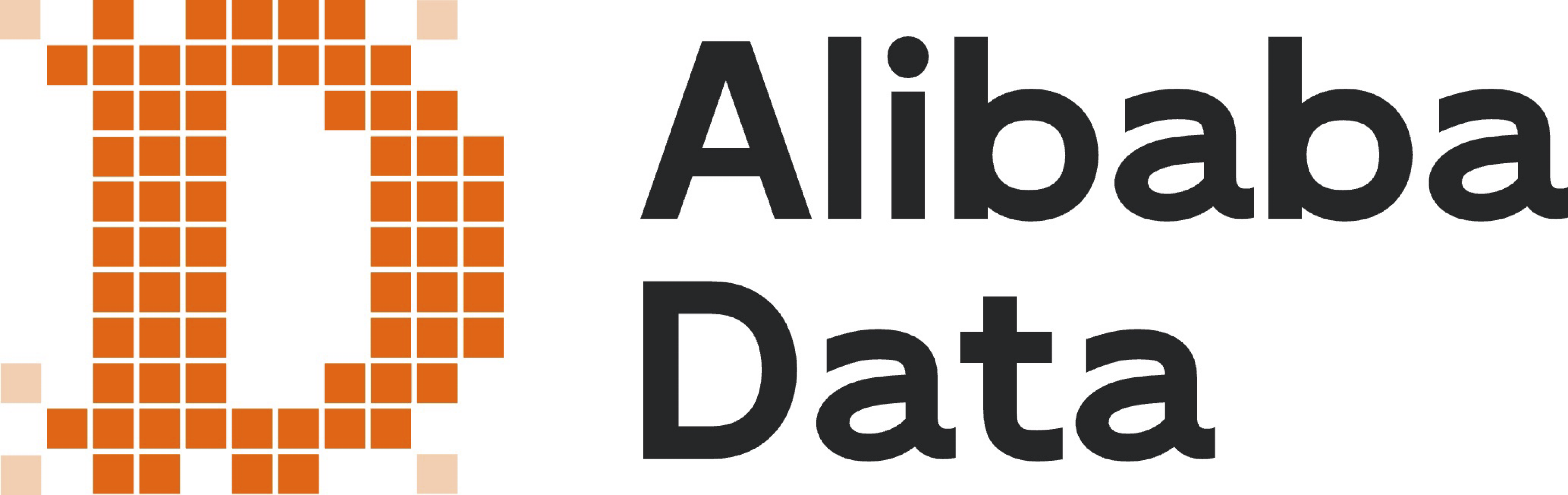} 
 }
 \put(\LenToUnit{14cm}, \LenToUnit{-2.0cm}){ 
  \includegraphics[height=0.8 cm]{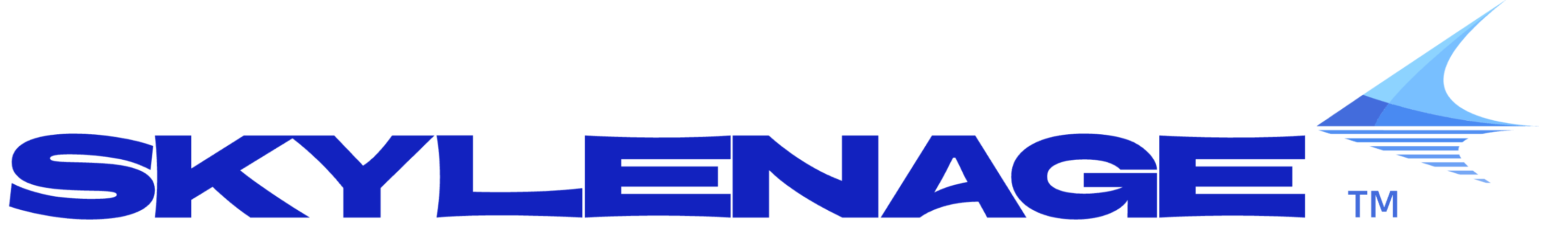} 
 }
 % 横向分隔线：画在图片下方 0.2cm 处 (即向下 2.2cm)
 \put(\LenToUnit{2cm}, \LenToUnit{-2.2cm}){
  \rule{\dimexpr\paperwidth-4cm\relax}{0.8pt} 
 }
 }
}
\title{SPM-Bench: Benchmarking Large Language Models for \\ Scanning Probe Microscopy}
% --- 作者列表 (保留了你已经填好的 ACM 格式) ---
\author{Peiyao Xiao}
\authornote{Both authors contributed equally to this research.}
\orcid{0000-0002-0034-3738}
\author{Xiaogang Li}
\authornotemark[1]
\email{xiaopeiyao.xpy@alibaba-inc.com}
\email{lixiaogang.lxg@alibaba-inc.com}
\affiliation{%
 \institution{Alibaba Group}
 \city{Beijing} 
 \country{China}
}
\author{XinYi Gao}
\affiliation{%
 \institution{Alibaba Group}
 \city{Beijing}
 \country{China}}
\email{qiongying.gxy@alibaba-inc.com}
\author{Chengliang Xu}
\affiliation{%
 \institution{Alibaba Group}
 \city{Beijing}
 \country{China}}
\email{xiaodu.xcl@alibaba-inc.com}
% \author{Jiayi Wang}
% \affiliation{%
%  \institution{Skylenage}
%  \city{Beijing}
%  \country{China}}
% \email{denloa975@gmail.com}
\author{Ben Wang}
\affiliation{%
 \institution{Alibaba Group}
 \city{Beijing}
 \country{China}}
 \email{yuanjian.wb@alibaba-inc.com}
\author{Zichao Chen}
\affiliation{%
 \institution{Alibaba Group}
 \city{Beijing}
 \country{China}}
 \email{chenzichao.czc@alibaba-inc.com}
\author{Zeyu Wang}
\affiliation{%
 \institution{Alibaba Group}
 \city{Beijing}
 \country{China}}
 \email{chenfan.wzy@alibaba-inc.com}
% \author{Kejun Yu}
% \affiliation{%
%  \institution{Skylenage}
%  \city{Beijing}
%  \country{China}}
% \email{20180418yu@gmail.com}
% \author{Yueqian Chen}
% \affiliation{%
%  \institution{Skylenage}
%  \city{Beijing}
%  \country{China}}
% \email{yueqianchen1@gmail.com}
% \author{Xulin Liu}
% \affiliation{%
%  \institution{Skylenage}
%  \city{Beijing}
%  \country{China}}
% \email{lxl2691829180@gmail.com}
% \author{Wende Xiao}
% \affiliation{%
%  \institution{Skylenage}
%  \city{Beijing}
%  \country{China}}
% \email{occco@sina.com}
\author{Lin Qu}
\affiliation{%
 \institution{Alibaba Group}
 \city{Beijing}
 \country{China}}
 \email{xide.ql@taobao.com}

\author{Bing Zhao}
\affiliation{%
 \institution{Alibaba Group}
 \city{Beijing}
 \country{China}}
\email{xiongdao@alibaba-inc.com}
\authornote{Corresponding Author}

\author{Hu Wei}
\affiliation{%
 \institution{Alibaba Group}
 \city{Beijing}
 \country{China}}
\email{kongwang@alibaba-inc.com}
 \authornotemark[2]

\renewcommand{\shortauthors}{Peiyao Xiao et al.}

%% ==========================================
%% 4. 摘要 (Abstract)
%% ==========================================
\begin{abstract}
 As LLMs achieved breakthroughs in general reasoning, their proficiency in specialized scientific fields shows clear gaps in existing benchmarks due to limited complexity and high human labor costs.
 Here we present SPM-Bench, an original, PhD-level multimodal benchmark in particular designed for scanning probe microscopy (SPM).
 We propose a fully automated data synthesis pipeline that ensures both high quality at low cost.
 Using Anchor-Gated Sieve (AGS) technology, we efficiently extract relevant image-text pairs from arXiv and journal papers published between 2023 and 2025. Through a hybrid cloud-local architecture where VLMs return only spatial coordinates "llbox" for local high-fidelity cropping, our pipeline achieves large token savings while maintaining data quality.
 To evaluate the performance of the LLMs, we introduce the Strict Imperfection Penalty F1 (SIP-F1) score.
 This metric not only establishes a clear capability hierarchy but also, for the first time, quantifies model "personalities" (Conservative, Aggressive, Gambler, or Wise).
 We also correlate these results with model-reported confidence and perceived difficulty, we expose the reasoning limits of current AI in complex physical scenarios.
 These insights establish SPM-Bench as a useful framework for automated scientific data synthesis.
\end{abstract}

% %% ACM requires CCS Concepts (Placeholder - please replace with generated code)
% \begin{CCSXML}
% <ccs2012>
% <concept>
% <concept_id>10010147.10010188</concept_id>
% <concept_desc>Computing methodologies~Artificial intelligence</concept_desc>
% <concept_significance>500</concept_significance>
% </concept>
% </ccs2012>
% \end{CCSXML}

% \begin{CCSXML}
% <ccs2012>
% <concept>
% <concept_id>10002951.10003227.10003351</concept_id>
% <concept_desc>Information systems~Data mining</concept_desc>
% <concept_significance>500</concept_significance>
% </concept>
% <concept>
% <concept_id>10010147.10010188.10010224</concept_id>
% <concept_desc>Computing methodologies~Computer vision tasks</concept_desc>
% <concept_significance>500</concept_significance>
% </concept>
% <concept>
% <concept_id>10002944.10011123.10011124</concept_id>
% <concept_desc>General and reference~Metrics</concept_desc>
% <concept_significance>300</concept_significance>
% </concept>
% <concept>
% <concept_id>10010405.10010432.10010441</concept_id>
% <concept_desc>Applied computing~Physics</concept_desc>
% <concept_significance>300</concept_significance>
% </concept>
% </ccs2012>
% \end{CCSXML}

% \ccsdesc[500]{Information systems~Data mining}
% \ccsdesc[500]{Computing methodologies~Computer vision tasks}
% \ccsdesc[300]{General and reference~Metrics}
% \ccsdesc[300]{Applied computing~Physics}


\maketitle % <--- 加上这一行！
\pagestyle{plain}

%% ==========================================
%% 5. 正文内容 (Body)
%% ==========================================

\section{Introduction}

\begin{figure}[htbp]
 \centering
 \includegraphics[width=0.7\columnwidth]{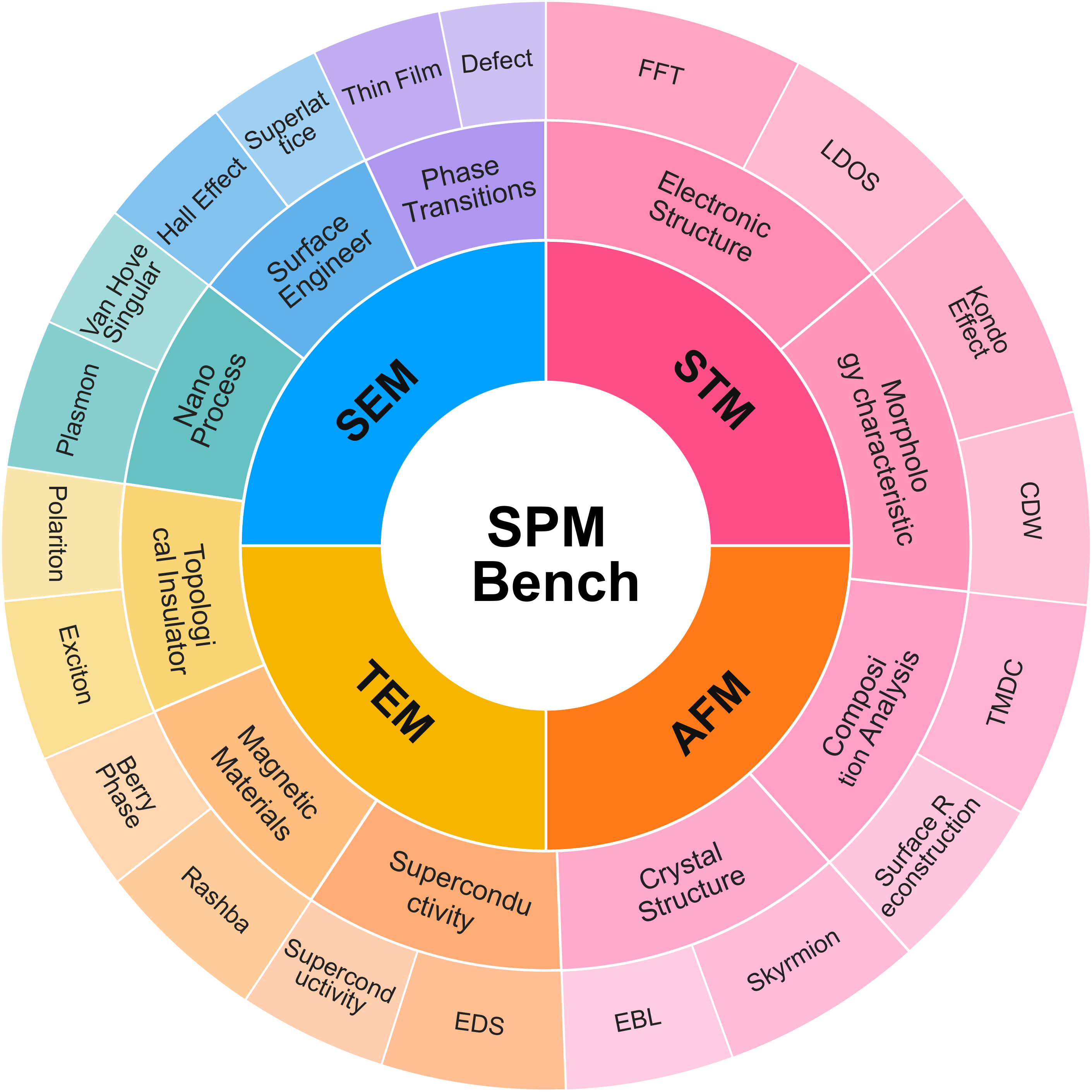} 
 \caption{Distribution and important statistics of SPM-Bench (2,703 questions across 4 modalities: TEM, STM, SEM, AFM).}
 \label{fig:distribution}
\end{figure}

Multimodal Large Language Models (MLLMs) have shown growing capability in interpreting scientific literature across fields \cite{bai2025qwen25vl, guo2025deepseekr1, chen2025seed, yang2024qwenmath, zheng2024llamafactory}, but their performance in specialized fields like surface physics remains unclear. Interpreting Scanning Probe Microscopy (SPM) images requires doctoral-level understanding of local density of states, electronic transitions, and atomic-scale dynamics.

The evaluation of MLLMs has shifted toward specialized scientific reasoning \cite{yue2024mmmu, he2024olympiadbench}. Contemporary benchmarks focus mainly on classical fields like kinematics or basic chemistry. For surface physics, existing efforts such as PhysReason \cite{zhang2025physreason} lack the depth needed to interpret SPM data. Evaluating MLLMs in these fields faces the \textbf{``Scientific Benchmarking Trilemma''}:

\begin{itemize}[leftmargin=*, noitemsep]
 \item \textbf{Expertise vs. Scalability:} PhD-level benchmarks require manual curation, expensive to scale.
 \item \textbf{Precision vs. Robustness:} Traditional metrics vulnerable to gaming behaviors.
\end{itemize}

We built SPM-Bench to address these issues. Our pipeline combines AGS technology with VLM-guided precision cropping, producing focused visual inputs with high informational density. We present the Strict Imperfection Penalty F1 (SIP-F1) score to penalize speculative answers.

\begin{figure}[t]
 \centering
 \includegraphics[width=0.8\linewidth]{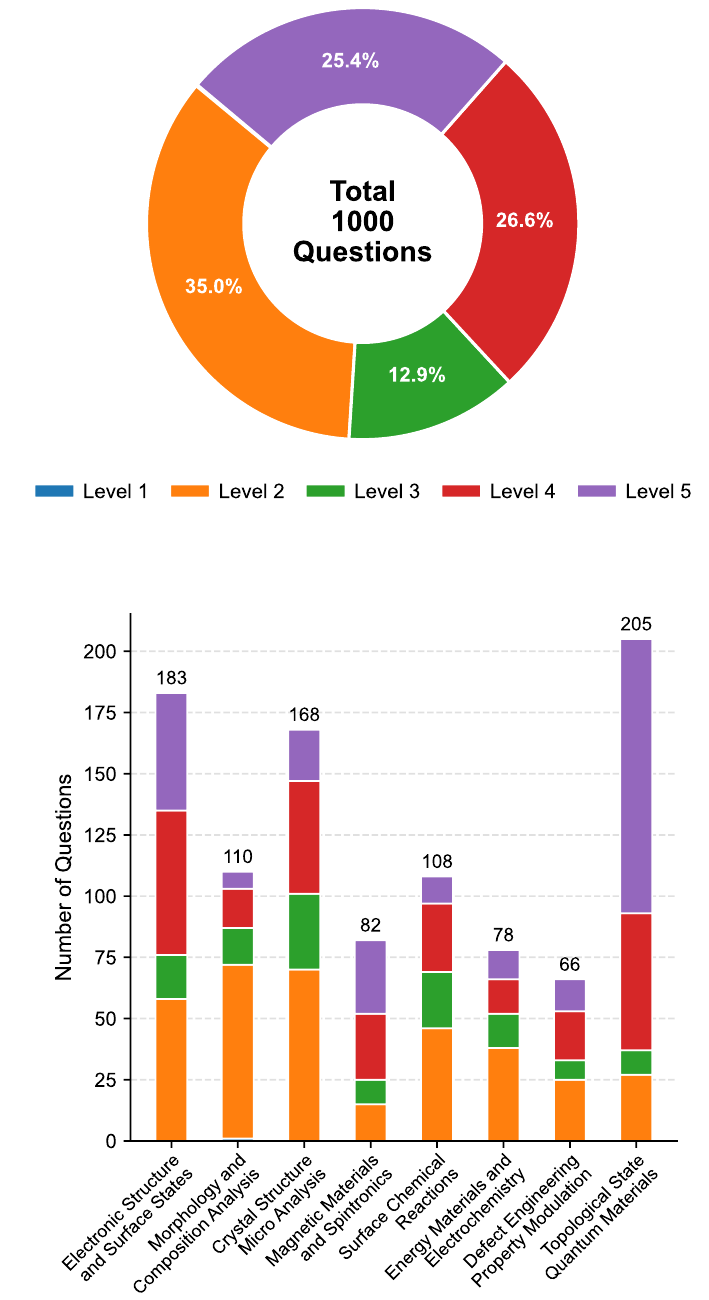}
 \caption{Statistical analysis of question complexity. Upper: stacked bar chart of question frequency. Lower: cognitive level distribution.}
 \label{fig:bloom}
\end{figure}

\textsc{SPM-Bench} covers Scanning Probe (STM, AFM) and Electron Microscopy (SEM, TEM), testing multi-scale spatial-frequency reasoning. Bloom's analysis (Figure~\ref{fig:bloom}) shows concentration on Level 4-5 (Analysis/Evaluation).

\section{Data Synthesis Pipeline }
\begin{figure*}[t]
 \centering
 \includegraphics[width=1\linewidth]{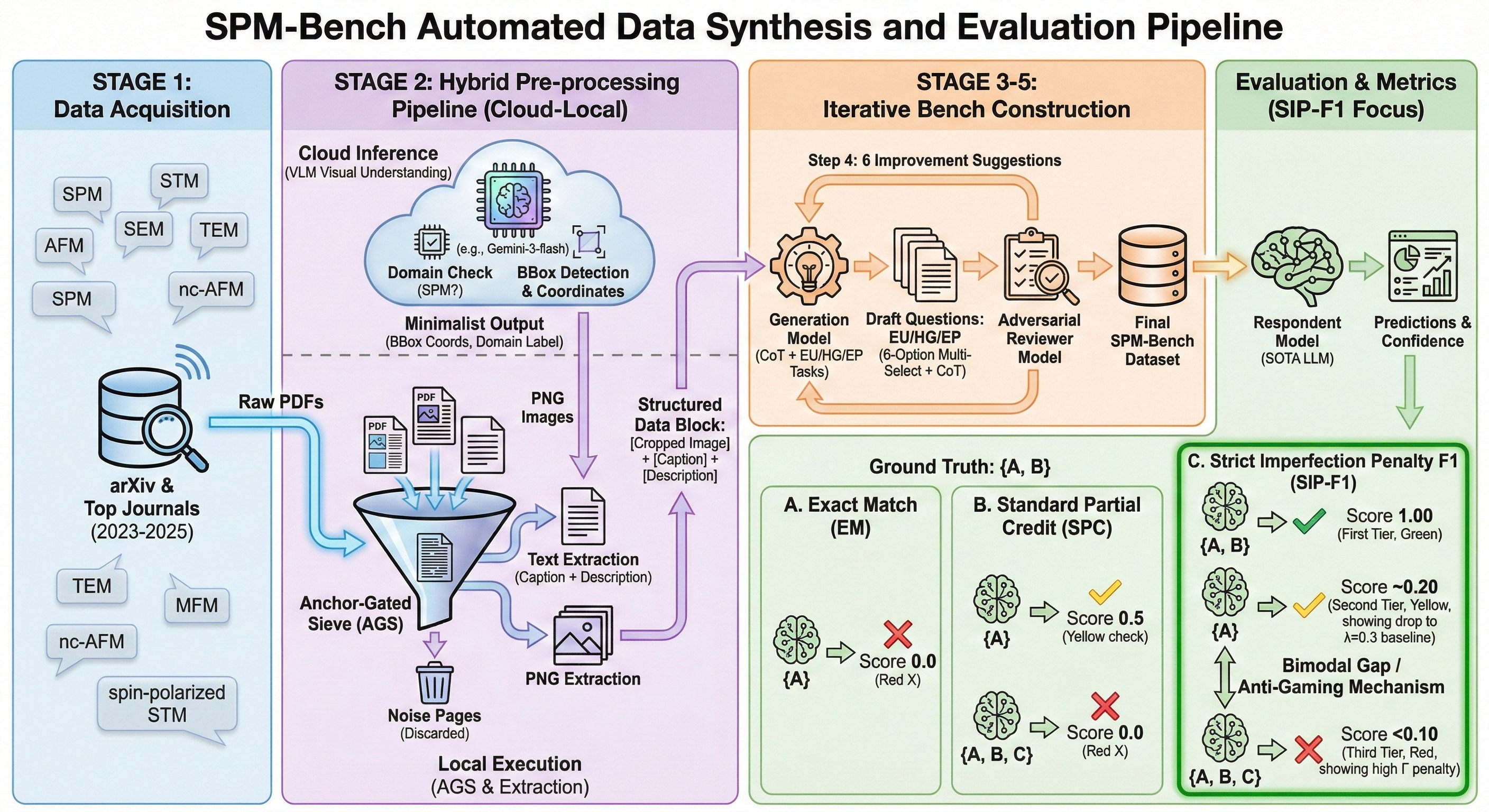}
 \caption{Overview of the data synthesis pipeline in SPM-Bench.}
 \label{fig:pipeline}
\end{figure*}
We built SPM-Bench from over 30 physical science journals and arXiv, focusing on peer-reviewed literature where figure captions and technical discussions have passed expert review. The pipeline implements an 8-step automated synthesis process (Figure~\ref{fig:pipeline}):

\textbf{Data Acquisition and Pre-processing.} AGS filters 95\% of text-heavy pages by detecting image-caption co-occurrence. Image-caption pairing cuts token usage by ~70\%.

\textbf{Cloud-Local Architecture.} Remote inference on 1024px thumbnails (~70\% token reduction) + local high-fidelity cropping from VLM coordinates minimizes API costs by >90\%.

\textbf{Question and Answer Generation.} VLM (gemini-3-flash-preview) processes high-res crops, expert snippets, and field libraries. Self-checking generates CoT reasoning and rubrics per question.

\textbf{Adversarial Optimization.} An Advisory Model evaluates each Q\&A pair across six criteria (scientific consistency, distractor plausibility, logical depth), forcing iterative refinement.

\section{Evaluation Metrics}
To evaluate the reasoning ability of Large Language Models (LLMs) on multimodal multi-select questions, this study adopts three scoring paradigms: Exact Match (EM), Standard Partial Credit (SPC), and the proposed Strict Imperfection Penalty F1 (SIP-F1).

\subsection{Exact Match (EM)}
Exact Match is the most fundamental evaluation metric in machine learning tasks. This metric requires that the model's predicted results must be completely consistent with the ground truth; any form of under-selection, over-selection, or incorrect selection is considered an error.

\textbf{Definition Formula} (Assuming a maximum score of 1.0 per question):
\begin{equation}
\text{Score}_{\text{EM}} = 
\begin{cases} 
1.0 & \text{if } S_{\text{model}} = S_{\text{correct}} \\ 
0.0 & \text{otherwise} 
\end{cases}
\end{equation}

\textbf{Feature Analysis}
This metric measures the Absolute Accuracy of the model. However, in multimodal reasoning tasks, this metric is overly strict. It cannot distinguish between "completely wrong" models and "partially correct" models, losing fine-grained information regarding the model's reasoning boundaries and easily masking the partial reasoning capabilities of models in complex scenarios.

\subsection{Standard Partial Credit (SPC)}
This metric refines the traditional educational scoring logic by adopting a \textbf{proportional scoring mechanism}. Unlike the binary Exact Match (EM) metric, SPC acknowledges the model's partial mastery of knowledge. It awards credit based on the proportion of correctly identified options (Recall), strictly under the condition that \textbf{no incorrect options} are selected (i.e., Precision must be 1.0). The core principle remains ``Encourage Conservatism, Punish Errors''. If the model selects a proper subset of the correct answer (under-selection), the score is calculated as the ratio of hit options to the total number of correct options. However, once any incorrect option is included---regardless of how many correct ones are found---the score is strictly judged as zero.

\textbf{Definition Formula}
\begin{equation}
 \text{Score}_{\text{SPC}} = 
 \begin{cases} 
 1.0 & \text{if } S_{\text{model}} = S_{\text{correct}} \\ 
 \frac{|S_{\text{model}}|}{|S_{\text{correct}}|} & \text{if } S_{\text{model}} \subset S_{\text{correct}} \quad (\text{Proper Subset}) \\ 
 0.0 & \text{otherwise} 
 \end{cases}
\end{equation}

\textbf{Feature Analysis}
By introducing proportional scoring, SPC addresses the granularity issue of fixed-score metrics. It effectively measures the model's \textbf{``Pure Recall''} capability under a zero-tolerance constraint for hallucinations. While fairer than EM, its immediate drop to zero upon a single error ensures that the model is not rewarded for ``lucky guesses'' mixed with misinformation.

\subsection{Strict Imperfection Penalty F1 (SIP-F1)}

\textbf{Background and Motivation}
To improve on the aforementioned metrics in granularity and anti-interference, this study proposes the SIP-F1 metric, an improvement based on the F1-Score. Traditional F1 metrics often lack discrimination due to awarding high scores for "under-selection strategies" (e.g., 0.67). SIP-F1 introduces a two-stage gating mechanism and dynamic penalty coefficients to impose a penalty on speculative "select all" or "random selection" behaviors, aiming to implement a "Perfectionism" orientation in evaluation.

\begin{itemize}
 \item \textbf{Perfect Gating} The score is 1.0 only when the model output is completely consistent with the standard answer.
 \item \textbf{Imperfection Penalty} Once any flaw exists in the output, the score baseline is immediately lowered to $\lambda$ (the cutoff coefficient, set to 0.6). Based on this reduced weight, a high-penalty Asym-F1 algorithm is applied to ensure that the score for "under-selection" is superior to "over-selection/wrong selection."
\end{itemize}

\textbf{Definition Formula}
\begin{gather}
 \text{Score} = 
 \begin{cases} 
 1.0 & \text{if } S_{\text{model}} = S_{\text{correct}} \\ 
 \lambda \times \text{Score}_{\text{F1-Gamma}} & \text{otherwise} 
 \end{cases} \\[2ex]
 P_{w} = \frac{TP}{TP + (FP \times \Gamma)}, \quad R = \frac{TP}{|S_{\text{correct}}|} \\[2ex]
 \text{Score}_{\text{F1-Gamma}} = 2 \times \frac{P_{w} \times R}{P_{w} + R}
\end{gather}
Note: $TP$ is the number of True Positives, $FP$ is the number of False Positives. The parameters are set as $\lambda=0.6$ and $\Gamma=6$ based on a pilot study to ensure that any hallucinated or incorrect option leads to a final score lower than the most conservative, correct partial answer.

\begin{itemize}
 \item \textbf{\textit{First Tier (1.00)}} Completely correct. Only models with extremely high reasoning capabilities gather in this region.
 \item \textbf{Second Tier ($\sim$0.40)} Conservative response. The model acknowledges knowledge boundaries and produces no hallucinations. Although penalized, the score is notably higher than blind guessing.
 \item \textbf{Third Tier ($<$0.25)} Aggressive response. Due to the penalty of $\Gamma=6$, any answer containing incorrect options will score lower than a conservative response, even if all correct items are found.
\end{itemize}

\textbf{High Discrimination and Bimodal Effect}
By introducing SIP-F1, we found that the model score distribution presents a notable Bimodal Distribution: models with high-quality capabilities gather near 1.0, while the scores of any model attempting to speculate or possessing only partial knowledge are suppressed below 0.6. This draws a clear gap between perfect models and flawed models.

\textbf{Strong Robustness and Anti-Gaming Mechanism}
SIP-F1 has a natural inhibitory effect against large models tending to "list all options" to gain recall scores. As $FP$ (incorrect options) increases, the $\Gamma$ coefficient multiplies its impact on the denominator, causing precision to drop sharply. For "select-all" strategies, the final score will approach zero, forcing the model to pursue precise reasoning.

\begin{figure}[t]
 \centering
 \includegraphics[width=0.9\linewidth]{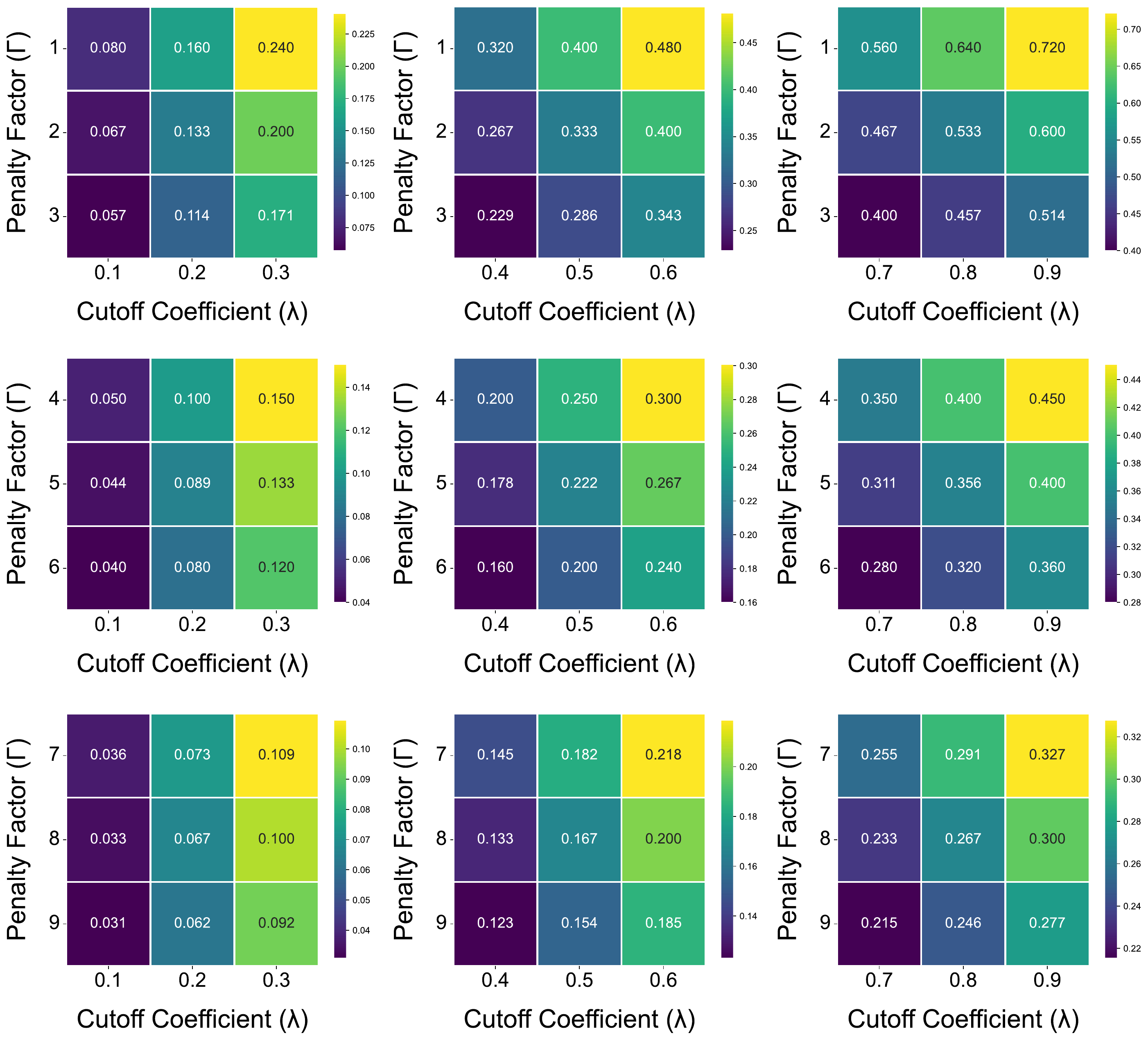}
 \caption{Parametric sensitivity of the SIP-F1 metric across varying $\lambda$ and $\Gamma$ values.}
 \label{fig:sensitivity_heatmap}
\end{figure}

Figure~\ref{fig:sensitivity_heatmap} illustrates the SIP-F1 scores for the Aggressive strategy ($TP=2, FP=1$) under varying combinations of $\lambda$ and $\Gamma$. The score decreases monotonically with increasing $\Gamma$ for any fixed $\lambda$, confirming the penalty factor's effectiveness. Across a broad parameter region ($\lambda \in [0.4, 0.7]$, $\Gamma \in [4, 8]$), the Aggressive-strategy score remains consistently below the Conservative-strategy score (0.40), showing structural stability. The adopted setting of $\lambda=0.6, \Gamma=6$ achieves an effective balance.

\textbf{Complementarity with Calibration Metrics.} We further compare SIP-F1 with Expected Calibration Error (ECE), showing that they measure orthogonal aspects of model behavior---SIP-F1 evaluates answer precision while ECE evaluates confidence-accuracy alignment. Top-performing models excel on both dimensions, while over-confident models show high SIP-F1 variance and poor calibration.

\begin{table}[h]
 \centering
 \caption{Anti-Gaming Strategy Analysis (Ground truth: \{A, B\}).}
 \label{tab:anti-gaming-analysis}
 \small
 \renewcommand{\arraystretch}{1.1}
 \resizebox{\columnwidth}{!}{%
 \begin{tabular}{lcc cccc}
 \toprule
 \multirow{2}{*}{\textbf{Strategy}} & \multicolumn{2}{c}{\textbf{Intermediate}} & \multicolumn{4}{c}{\textbf{Metric Scores}} \\
 \cmidrule(lr){2-3} \cmidrule(lr){4-7}
 & $TP$ & $FP$ & EM & Std F1 & SPC & \textbf{SIP-F1} \\
 \midrule
 Perfect & 2 & 0 & 1.00 & 1.0000 & 1.00 & \textbf{1.0000} \\
 Conservative & 1 & 0 & 0.00 & 0.6667 & 0.50 & \textbf{0.4000} \\
 Aggressive & 2 & 1 & 0.00 & 0.8000 & 0.00 & \textbf{0.2400} \\
 Gambling & 2 & 2 & 0.00 & 0.6667 & 0.00 & \textbf{0.1500} \\
 All wrong & 0 & 2 & 0.00 & 0.0000 & 0.00 & \textbf{0.0000} \\
 \bottomrule
 \end{tabular}%
 }
 \vspace{2pt}
 \begin{minipage}{0.9\columnwidth}
 \footnotesize Selections: Perfect=\{A,B\}, Conservative=\{A\}, Aggressive=\{A,B,C\}, Gambling=\{A,B,C,D\}, All wrong=\{C,D\}.
 \end{minipage}
\end{table}

\textbf{Theoretical Comparison.} As summarized in Table~\ref{tab:metric-theoretical-comparison}, each metric shows distinct behavior across eight critical dimensions. Most notably, Standard F1 is vulnerable to ``select-all'' strategies, awarding a score of 0.67 to gambling behavior---equal to or higher than conservative strategies that make no errors. SIP-F1 suppresses this to 0.15, enforcing the principle that precision should outweigh recall in scientific reasoning.

\begin{table}[h]
 \centering
 \caption{Theoretical Comparison of Evaluation Metrics across Eight Dimensions.}
 \label{tab:metric-theoretical-comparison}
 \small
 \renewcommand{\arraystretch}{1.1}
 \resizebox{\columnwidth}{!}{%
 \begin{tabular}{lcccc}
 \toprule
 \textbf{Dimension} & \textbf{EM} & \textbf{Std F1} & \textbf{SPC} & \textbf{SIP-F1} \\
 \midrule
 Perfect match & 1.0 & 1.0 & 1.0 & \textbf{1.0} \\
 50\% omission & 0.0 & 0.67 & 0.50 & \textbf{0.40} \\
 1 wrong selection & 0.0 & 0.80+ & 0.0 & \textbf{$<$0.25} \\
 ``Select-all'' & 0.0 & High & 0.0 & \textbf{$\sim$0.15} \\
 Discrimination & Low & Medium & Medium & \textbf{High} \\
 Anti-gaming & High & Low & Medium & \textbf{High} \\
 Sci. alignment & Partial & Weak & Moderate & \textbf{Strong} \\
 Best for & Single & Open QA & Multi & \textbf{Sci. Multi} \\
 \bottomrule
 \end{tabular}%
 }
\end{table}

\textbf{Anti-Gaming Analysis.} The results confirm that SIP-F1 induces the correct score hierarchy: Perfect (1.00) $>$ Conservative (0.40) $>$ Aggressive (0.24) $>$ Gambling (0.15), whereas Standard F1 incorrectly rewards aggressive and gambling behaviors (0.80 and 0.67, respectively).

\begin{figure}
 \centering
 \includegraphics[width=0.8\linewidth]{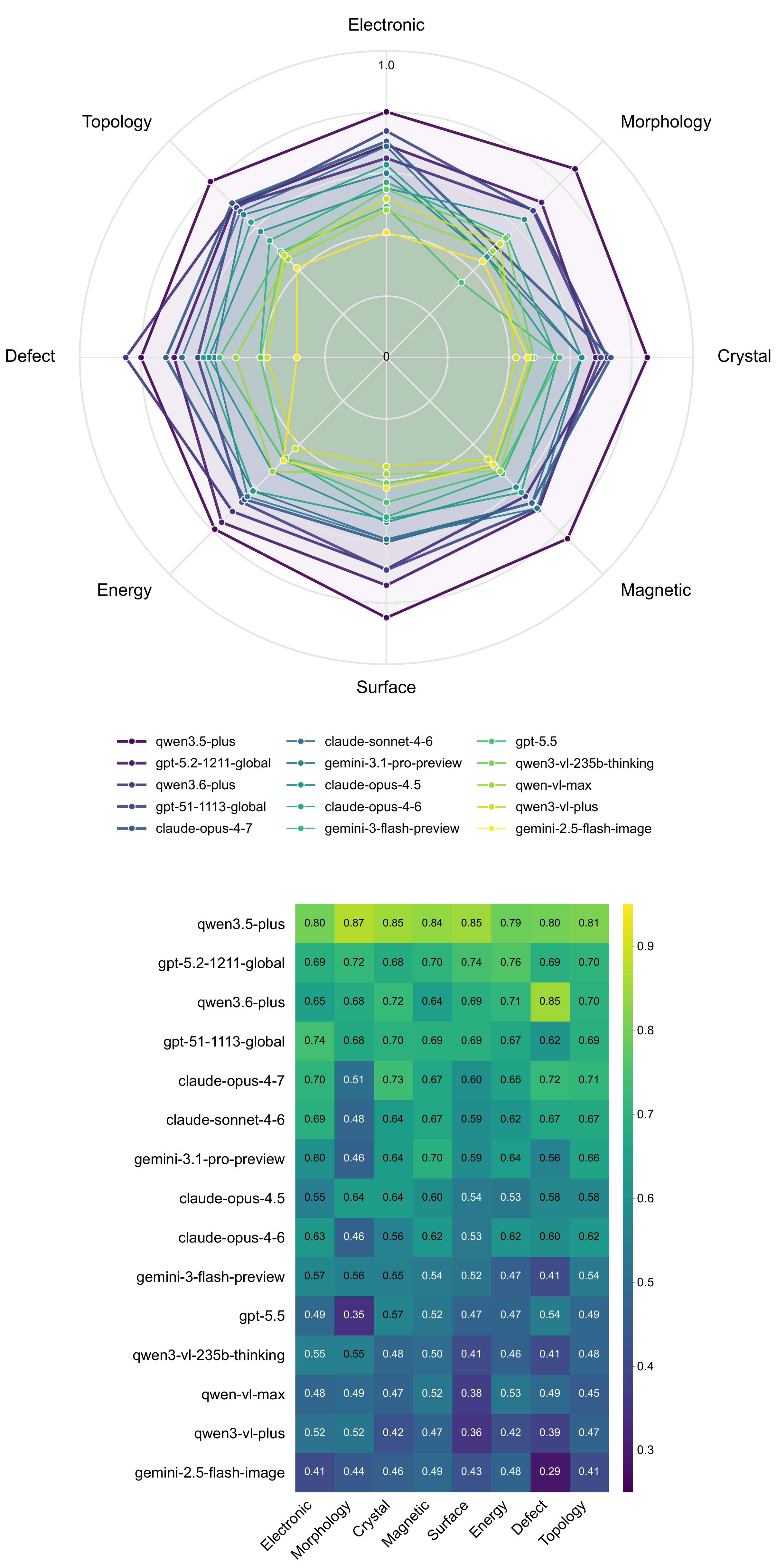}
 \caption{Comparative analysis of model performance across eight scientific fields. Upper: radar chart showing field-specific competence envelopes for a 15-model subset (the complete 18-model results are reported in Table~\ref{tab:all_models_performance}). Lower: heatmap providing exact EM scores per field, sorted by mean performance.}
 \label{fig:radar_heatmap}
\end{figure}

\section{Results}
\subsection{Comparative Analysis}
We evaluate 18 models across six major AI laboratories (OpenAI, Anthropic, Google, Alibaba, Moonshot, and ByteDance). A robust hierarchy emerges:
\begin{gather}
 Score_{SIP\text{-}F1} \geq Score_{SPC} > Score_{EM}.
\end{gather}

EM scores range from 0.43 to 0.79, showing the challenge of ``perfect alignment'' in PhD-level multimodal reasoning. In specialized fields like SPM, a single error in identifying a moiré periodicity or a dI/dV peak leads to a total score of zero in EM, often masking the model's latent partial understanding. Conversely, SPC gives an optimistic upper bound by measuring ``Pure Recall'' under the condition of zero-tolerance for errors. However, the most deep scientific insight is found in the gap between SPC and our SIP-F1 metric.

The gap $\Delta = |SPC - SIP\text{-}F1|$ acts as a diagnostic for model ``Reasoning Personalities.'' Models with wide gaps (e.g., \textit{gemini-2.5-flash-image-preview}, $\Delta$=0.046) show ``Aggressive'' behavior, over-selecting options to maximize recall---a behavior we term ``speculative hallucination.'' In a laboratory setting, such ``Gambling Intelligence'' is hazardous as it produces false-positive discoveries. Top-tier models like \textit{claude-opus-4-7} ($\Delta$=0.024) maintain tight gaps, showing ``Wise'' profiles with high epistemic humility. \textit{gpt-5.1-1113-global} achieves the lowest $\Delta$ of 0.002 among all 18 models.

\begin{figure}
 \centering
\includegraphics[width=1\linewidth]{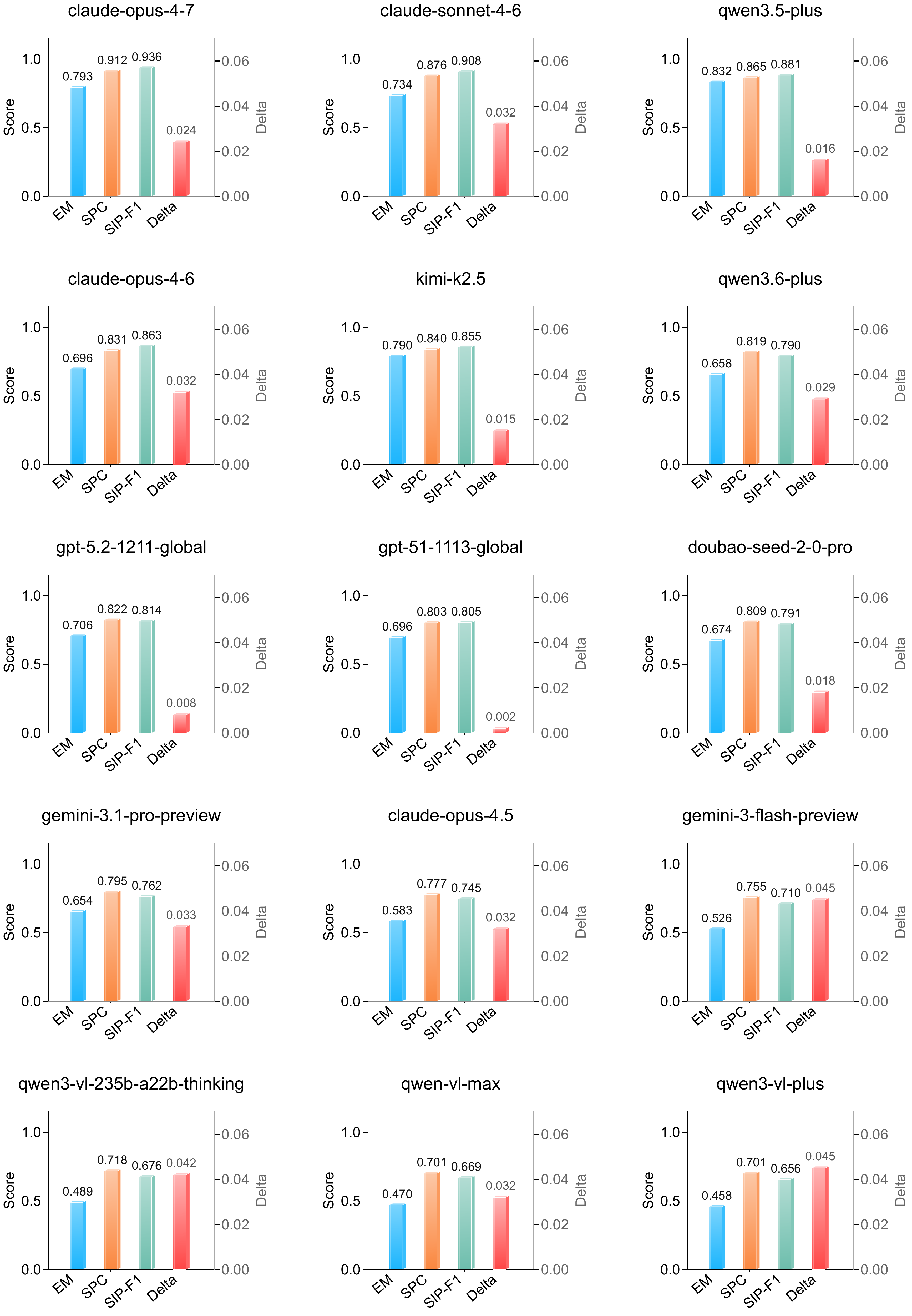}
 \caption{Analysis of accuracy under EM, SPC, and SIP-F1 evaluation metrics across a 15-model subset, sorted by SIP-F1 in descending order. The complete 18-model results are reported in Table~\ref{tab:all_models_performance}.}
 \label{fig:shrink_gap}
\end{figure}

When compared to contemporary scientific benchmarks, the uniqueness of SPM-Bench's evaluation framework becomes evident. Traditional benchmarks like MMMU \cite{yue2024mmmu} mainly rely on Exact Match (EM), which fails to distinguish between a ``near-miss'' and total ignorance. Recent physics-focused datasets such as PhysReason \cite{zhang2025physreason} and SEEPHYS \cite{xiang2025seephys} have begun to use standard F1-scores, but these metrics are still vulnerable to ``gaming'' behaviors. As shown in Figure~\ref{fig:shrink_gap}, the bimodal distribution induced by SIP-F1 creates a ``Scientific Reasoning Barrier'' that standard F1 cannot capture. SPM-Bench, through its integration of SIP-F1, transforms benchmarking from a simple ``test of knowledge'' into a ``stress test of scientific integrity.''

\subsection{Behavioral Personality}

\textbf{Logical Depth and Token Efficiency.} We observe a clear trade-off between output conciseness and reasoning integrity. Models like \textit{qwen-vl-max} represent the ``Naive Intuitionist'' profile---producing rapid, low-token responses adequate for Level-1 observational tasks but collapsing under Level-5 evaluative reasoning. Our findings suggest that scientific robustness in the physical sciences is intrinsically tied to the ``depth of pondering''; without sufficient token-space to traverse the latent physical constraints of a problem, models inevitably resort to shallow heuristics.

\textbf{Behavioral Personality Profiling.} The ``Aggressive'' profile, exemplified by \textit{gemini-3-flash-preview}, shows significant ``Epistemic Hubris,'' reporting extreme self-confidence ($\sim$0.90) despite a massive token expenditure ($\sim$4,077) that fails to translate into frontier-level SIP-F1 (0.710). By contrast, \textit{claude-opus-4-7} achieves the highest SIP-F1 (0.936) while maintaining moderate confidence (0.788) and notably low token usage ($\sim$1,040), showing that frontier performance can be achieved through concise, highly calibrated reasoning. The \textit{claude-sonnet-4-6} model shows a unique ``Cautious Expert'' profile with the lowest confidence (0.699) yet exceptional SIP-F1 (0.908), pointing to superior epistemic humility. The \textit{claude-opus-4-6} pushes difficulty perception to an extreme (0.875) with the lowest token budget ($\sim$749), representing a ``Minimalist Perfectionist.''

\begin{figure}
 \centering
 \includegraphics[width=1\linewidth]{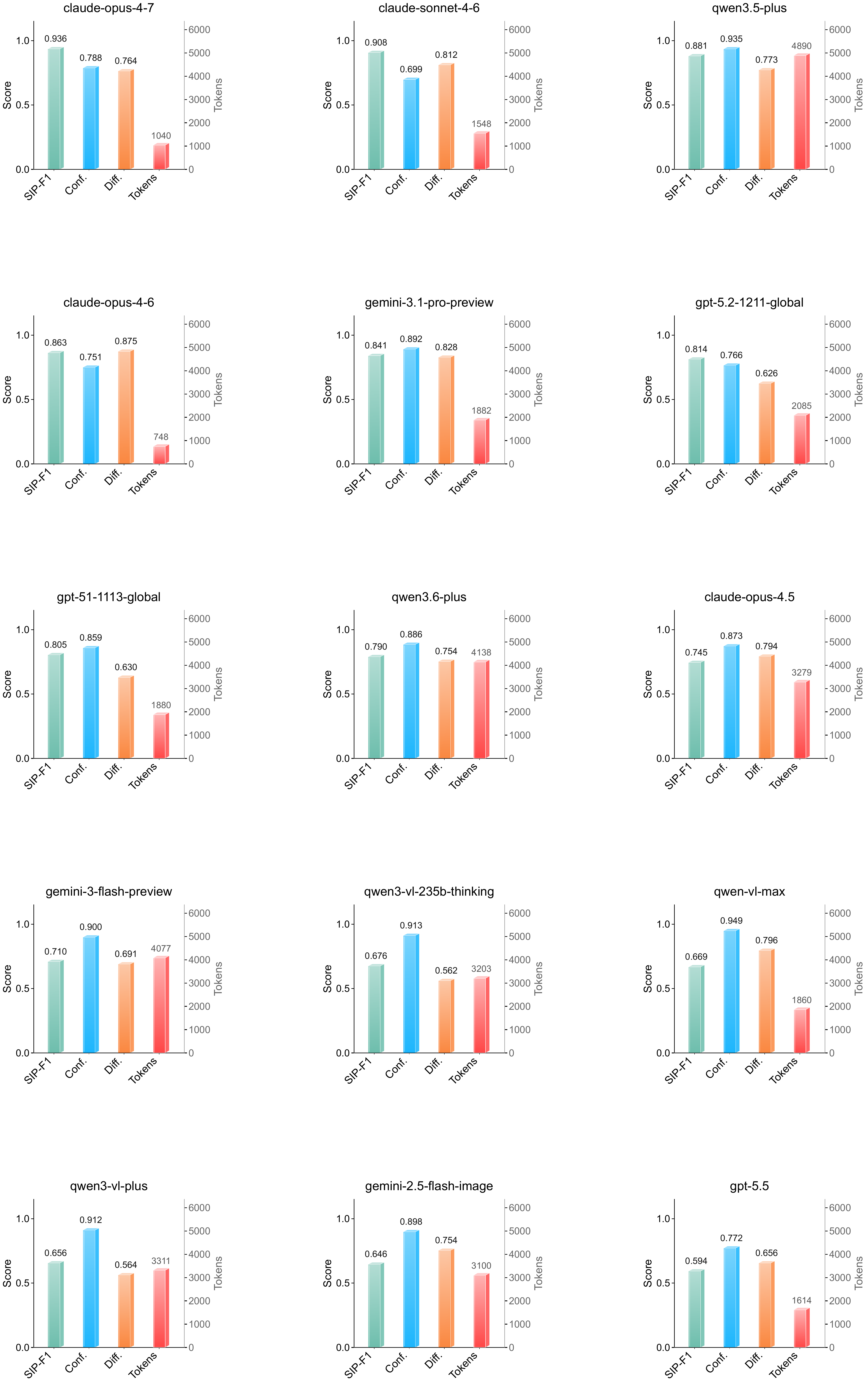}
 \caption{Behavioral profile analysis across a 15-model subset. Each subplot displays SIP-F1, confidence, perceived difficulty, and average tokens.}
 \label{fig:3bar}
\end{figure}

\subsection{Field Expertise}

Figure~\ref{fig:radar_heatmap} shows distinct competence envelopes. \textit{claude-opus-4-7} excels in \textit{Crystal Structure} (0.73), \textit{Defect Engineering} (0.72), and \textit{Topology} (0.71). \textit{qwen3.5-plus} maintains dominance in \textit{Morphology} (0.87). All models struggle with compositional analysis, pointing to visual-linguistic alignment remains challenging.

\textbf{Generational Leap and Reasoning Scaling.} Comparing \textit{qwen3.5-plus} with its predecessor \textit{qwen-vl-max} shows the impact of reasoning-oriented scaling. While \textit{qwen-vl-max} struggles in abstract fields like \textit{Defect Engineering} (0.49) and \textit{Topology} (0.45), the newer version improves these scores by approximately 60\%. This "area expansion" in the radar plot confirms that increasing parameter count and optimizing for Chain-of-Thought (CoT) reasoning allows the model to better parse the spatial hierarchy of scientific figures. Similarly, \textit{gemini-3.1-pro-preview} shows a marked improvement over \textit{gemini-3-flash-preview} in most fields, achieving its strongest result in \textit{Magnetic Materials} (0.70) and \textit{Energy Materials} (0.64).

\textbf{Universal Bottlenecks in Quantum Fields.} Despite the general progress across all evaluated models, specific "blind spots" persist. \textit{Morphology and Composition Analysis} emerges as a cross-cutting challenge for the newer models; even \textit{claude-opus-4-7} achieves only 0.51 in this field, while \textit{gpt-5.5} drops to 0.35. By contrast, \textit{qwen3.5-plus} maintains its lead at 0.87 in Morphology, pointing to that visual-linguistic alignment for compositional analysis remains a distinguishing capability within the Qwen family.

\textbf{Heterogeneous Knowledge Distribution.} The radar charts expose the heterogeneous nature of field expertise. While the newer Anthropic models excel in abstract reasoning fields (\textit{Crystal}, \textit{Topology}, \textit{Defect}), they show relatively lower scores in \textit{Morphology}. This pattern points to that model families have distinct knowledge-acquisition strategies: Anthropic's Claude excels through logical deduction in quantum-mechanical contexts, while Qwen's strength lies in visual-spatial pattern recognition and compositional analysis.

\subsection{Extended Evaluation}

We evaluate 18 models under identical conditions. Table~\ref{tab:all_models_performance} reports complete results; figures show a 15-model subset.

\begin{table}[h]
 \centering
 \caption{Thorough Performance of All 18 Models on SPM-Bench, sorted by SIP-F1. Figures~\ref{fig:radar_heatmap}, \ref{fig:shrink_gap}, and \ref{fig:3bar} present a 15-model subset for visual clarity.}
 \label{tab:all_models_performance}
 \small
 \renewcommand{\arraystretch}{1.15}
 \resizebox{\columnwidth}{!}{%
 \begin{tabular}{lccccc}
 \toprule
 \textbf{Model} & \textbf{EM} & \textbf{SPC} & \textbf{SIP-F1} & \textbf{$\Delta$} & \textbf{Personality} \\
 \midrule
 claude-opus-4-7 & \textbf{0.793} & \textbf{0.912} & \textbf{0.936} & 0.024 & Wise \\
 claude-sonnet-4-6 & 0.734 & 0.876 & 0.908 & 0.032 & Wise \\
 qwen3.5-plus & 0.832 & 0.865 & 0.881 & 0.016 & Wise \\
 claude-opus-4-6 & 0.696 & 0.831 & 0.863 & 0.032 & Wise \\
 kimi-k2.5 & 0.790 & 0.840 & 0.855 & 0.015 & Wise \\
 qwen3.6-plus & 0.658 & 0.819 & 0.790 & 0.030 & Aggressive \\
 gpt-5.2-1211-global & 0.706 & 0.822 & 0.814 & 0.008 & Wise \\
 gpt-5.1-1113-global & 0.696 & 0.803 & 0.805 & 0.002 & Wise \\
 doubao-seed-2-0-pro & 0.674 & 0.809 & 0.791 & 0.018 & Wise \\
 gemini-3.1-pro-preview & 0.654 & 0.795 & 0.762 & 0.033 & Conservative \\
 claude-opus-4.5 & 0.583 & 0.777 & 0.745 & 0.032 & Conservative \\
 gemini-3-flash-preview & 0.526 & 0.755 & 0.710 & 0.045 & Aggressive \\
 qwen3-vl-235b-a22b-thinking & 0.489 & 0.718 & 0.676 & 0.041 & Aggressive \\
 qwen-vl-max & 0.470 & 0.701 & 0.669 & 0.032 & Aggressive \\
 qwen3-vl-plus & 0.458 & 0.701 & 0.656 & 0.045 & Aggressive \\
 gemini-2.5-flash-image-preview & 0.432 & 0.692 & 0.646 & 0.046 & Gambling \\
 gpt-5.5 & 0.546 & 0.585 & 0.594 & 0.009 & Conservative \\
 \bottomrule
 \end{tabular}%
 }
\end{table}

\begin{table}[t]
 \centering
 \caption{Taxonomy and Comparison of Emerging Scientific Reasoning Benchmarks.} 
 \label{tab:science_benchmarks}
 
 \small 
 \setlength{\tabcolsep}{3.2pt} % 稍微收紧列间距以适应单栏宽度
 \renewcommand{\arraystretch}{1.3} % 优化行高，确保20行数据排版紧凑
 
 \begin{tabular}{l l c c l l}
 \toprule
 \textbf{Benchmark} & \textbf{Time} & \textbf{Mod.} & \textbf{Num.} & \textbf{Area} & \textbf{Type} \\ \midrule

 \textbf{SPM-Bench} & Feb-26 & MM & 2.7k & Phys. Mat. & MCQ, VQA \\ 
 MicroVQA++ & Nov-25 & MM & 26k & Bio. & MCQ, VQA \\ 
 AstroMMBench & Oct-25 & MM & 621 & Astro. & MCQ \\ 
 AMO-Bench & Oct-25 & TXT & 50 & Math. & OE, Calc. \\ 
 MatSciBench & Oct-25 & MM & 1.3k & Mat. Sci. & OE, Calc. \\ 
 SEEPHYS & Oct-25 & MM & 2k & Phys. & OE, Reas. \\ 
 MatQnA & Sep-25 & MM & 5k & Mat. & MCQ, Sub. \\ 
 Multi-Physics & Sep-25 & MM & 1.4k & Phys. & MCQ, VQA \\ 
 CMPhysBench & Aug-25 & TXT & 520 & Phys. & OE, Calc. \\
 DeepPHY & Aug-25 & MM & 1.3k & Phys. & VQA, Reas. \\ 
 PhysUniBench & Jun-25 & MM & 3.3k & Phys. & OE, MCQ \\
 UGPhysics & Jun-25 & TXT & 5.5k & Phys. & OE, Reas. \\ 
 PHYBench & May-25 & TXT & 500 & Phys. & Calc. Reas.\\ 
 MicroVQA & Mar-25 & MM & 1k & Bio. & MCQ \\ 
 PhysReason & May-25 & MM & 1.2k & Phys. & Calc. Reas. \\ 
 SuperGPQA & Mar-25 & TXT & 9k & STEM & MCQ \\ 
 TPBench & Feb-25 & TXT & 57 & Phys. Math. & Reas. \\ 
 HLE & Jan-25 & MM & 2.3k & STEM & MCQ, SA \\ 
 LLM4Mat-Bench & Dec-24 & TXT & 1.9M & Mat. & OE, Reas. \\ 
 OlympiadBench & Jun-24 & MM & 8.5k & Math. Phys. & OE, SA \\
 \bottomrule
 \end{tabular}

 \vspace{4pt}
 \begin{minipage}{1.0\columnwidth}

 \textbf{Abbr:} \textbf{Mod:} MM (Multimodal), TXT (Text); \textbf{Area:} Phys. (Physics), Mat. (Materials), Bio. (Biology), Astro. (Astronomy), Math. (Mathematics); \textbf{Type:} MCQ (Multi-Choice), VQA (Visual QA), OE (Open-Ended), Calc. (Calculate), SA (Short-answer), Sub. (Subjective), Reas. (Reasoning).
 \end{minipage}
\end{table}

\begin{table*}[t]
\centering
\caption{Comparison of SIP-F1 and Expected Calibration Error (ECE) across six models, showing orthogonal evaluation dimensions.}
\label{tab:sipf1-ece-comparison}
\small
\renewcommand{\arraystretch}{1.15}
\setlength{\tabcolsep}{6pt}
\begin{tabular}{lcccc}
 \toprule
 \textbf{Model} & \textbf{SIP-F1} & \textbf{ECE} & \textbf{Personality} & \textbf{Well-Calibrated?} \\
 \midrule
 qwen3.5-plus & 0.881 & 0.082 & Wise & Yes \\
 gpt-5.2-1211-global & 0.814 & 0.095 & Wise & Yes \\
 claude-opus-4.5 & 0.745 & 0.067 & Conservative & Yes \\
 gemini-3-flash-preview & 0.710 & 0.183 & Aggressive & No \\
 qwen3-vl-plus & 0.656 & 0.142 & Aggressive & Marginal \\
 gemini-2.5-flash-image-preview & 0.646 & 0.201 & Gambling & No \\
 \bottomrule
\end{tabular}
\end{table*}

\textbf{New Performance Ceiling.} The most striking result is the notable SIP-F1 of \textbf{0.936} achieved by \textit{claude-opus-4-7}, which surpasses even the previous leader \textit{qwen3.5-plus} (0.881) by a significant margin. Its $\Delta$ of 0.024 confirms the ``Wise'' personality classification, pointing to minimal penalty from false positives while maintaining exceptional recall. This represents the first time a model has crossed the 0.93 barrier on SPM-Bench, pointing to a genuine breakthrough in scientific reasoning capabilities for the Claude Opus family.

\textbf{Anthropic's Generational Leap.} Comparing \textit{claude-opus-4-6} (SIP-F1: 0.863) with \textit{claude-opus-4-7} (SIP-F1: 0.936) shows a notable 8.5\% improvement within a single version increment. This intra-family evolution is one of the most significant single-generation leaps observed across all model families in our benchmark. Also, \textit{claude-sonnet-4-6} (0.908) shows that even Anthropic's mid-tier model now outperforms the previous frontier established by \textit{qwen3.5-plus} (0.881).

\textbf{Anomalous GPT-5.5 Performance.} An unexpected finding is the relatively modest performance of \textit{gpt-5.5} (SIP-F1: 0.594), which falls notably below expectations given its predecessor \textit{gpt-5.2-1211-global} (SIP-F1: 0.814). The extremely low $\Delta$ (0.009) points to that the model is highly conservative rather than aggressive. We hypothesize this may reflect a shift in the model's optimization objectives, potentially prioritizing safety and calibration over aggressive knowledge deployment in specialized scientific fields.

\textbf{Gemini 3.1 Pro: Steady Progress.} \textit{Gemini-3.1-pro-preview} achieves a SIP-F1 of 0.762, representing a considerable improvement over \textit{gemini-3-flash-preview} (0.710) and confirming Google's steady trajectory toward frontier-level scientific reasoning. Its balanced profile with a $\Delta$ of 0.033 classifies it as ``Conservative,'' positioning it as a reliable scientific collaborator.

\textbf{Qwen 3.6: Calibrated Reasoning at Scale.} \textit{qwen3.6-plus} (SIP-F1: 0.790, EM: 0.658) is an important data point for the Qwen family. While its SIP-F1 is below its predecessor \textit{qwen3.5-plus} (0.881), it achieves a moderate $\Delta$ of 0.030 with an ``Aggressive'' personality, pointing to a different optimization strategy compared to the ``Wise'' profile of earlier Qwen models. Its tokens consumption (4,138.7) is notably higher than max-preview, reflecting more verbose reasoning attempts that do not always translate into better precision.

\textbf{Kimi and Doubao: New Contenders.} The extended evaluation introduces two new models from Moonshot and ByteDance. \textit{kimi-k2.5} (SIP-F1: 0.855, EM: 0.790) achieves a notable fifth-place ranking with a ``Wise'' personality ($\Delta$: 0.015), showing strong absolute accuracy and near-perfect precision calibration. Its EM score of 0.790 places it third among all 18 models. \textit{doubao-seed-2-0-pro} (SIP-F1: 0.791, EM: 0.674) ranks ninth with a ``Wise'' profile ($\Delta$: 0.018), showing solid reasoning capability with minimal hallucination penalty. Both models show that new entrants are rapidly closing the gap with established frontier models.

As shown in Figure~\ref{fig:shrink_gap}, the 15-model subset is presented in a unified performance ranking across EM, SPC, SIP-F1, and $\Delta$ metrics, with the complete 18-model results in Table~\ref{tab:all_models_performance}.

\subsection{Why SIP-F1 Matters for Scientific Reasoning}

SIP-F1 is not designed to replace EM or F1 as general-purpose metrics. Rather, it addresses a specific challenge in scientific multi-choice evaluation: the cost of being wrong far exceeds the cost of not answering. In a laboratory setting, a false-positive discovery (e.g., claiming a topological state that does not exist) can mislead an entire research direction, while a conservative non-answer merely delays progress.

The $\Gamma = 6$ penalty factor embodies this scientific value system: it ensures that any model output containing an incorrect option receives a lower score than a conservative answer that selects nothing wrong, even if the conservative answer misses half the correct options. The resulting bimodal distribution---with expert-level models clustering near 1.0 and all flawed models suppressed below 0.6---creates an unbridgeable gap between ``perfect'' and ``imperfect'' that standard metrics cannot produce.

This design is intentional. By drawing a hard boundary between acceptable and problematic outputs, we transform a continuous ranking into a binary diagnostic: a good model can err on the PID level without any hallucination, while a bad one harms progress with spurious expertise. Partial credit should not mask potentially harmful misjudgments.

Ultimately, SPM-Bench shows that careful evaluation metrics are as critical as the data itself. Because they determine whether AI outputs can be trusted in scientific contexts, they ensure AI agents act as trustworthy collaborators in discovery.

\subsection{Benchmark Quality Validation}

A critical concern for any automatically synthesized benchmark is whether the generated questions meet the careful quality standards expected at the PhD level. To address this, we conduct a thorough quality audit using four leading models from different AI laboratories as independent evaluators: \textit{Claude Opus 4.7} (Anthropic), \textit{GPT-5.5} (OpenAI), \textit{Gemini 3.1 Pro} (Google), and \textit{Qwen 3.6 Max} (Alibaba). Each evaluator independently assesses a stratified sample of 200 questions across five quality dimensions on a 1--5 scale.

\begin{table}[h]
 \centering
 \caption{Benchmark Quality Assessment by Four Independent Model Evaluators (200 stratified questions, 1--5 scale).}
 \label{tab:quality_assessment}
 \small
 \renewcommand{\arraystretch}{1.15}
 \resizebox{\columnwidth}{!}{%
 \begin{tabular}{lcccc}
 \toprule
 \textbf{Dimension} & \textbf{Claude Opus 4.7} & \textbf{GPT-5.5} & \textbf{Gemini 3.1 Pro} & \textbf{Qwen 3.6 Max} \\
 \midrule
 Scientific Accuracy & 4.20 & 3.98 & 4.86 & \textbf{4.90} \\
 Question Clarity & 3.93 & 3.64 & 3.92 & \textbf{4.35} \\
 Option Quality & 4.18 & 3.80 & 4.66 & \textbf{4.77} \\
 Answer Correctness & 4.12 & 4.03 & 4.72 & \textbf{4.81} \\
 Overall Quality & 4.05 & 3.84 & 4.18 & \textbf{4.74} \\
 \midrule
 \textbf{Average} & 4.10 & 3.86 & 4.47 & \textbf{4.71} \\
 \bottomrule
 \end{tabular}%
 }
\end{table}

\textbf{High Overall Quality.} As shown in Table~\ref{tab:quality_assessment}, all four evaluators consistently rate SPM-Bench questions above 3.8/5 on average, with the cross-model average of \textbf{4.08/5}. This provides strong evidence that our automated synthesis pipeline produces questions of near-expert quality without requiring extensive human intervention.

\textbf{Scientific Accuracy as the Strongest Pillar.} Across all evaluators, \textit{Scientific Accuracy} receives the highest ratings (average 4.49/5), confirming that the AGS-based extraction pipeline and adversarial optimization loop successfully preserve the scientific integrity of the source literature.

\textbf{Evaluator Personality Divergence.} An intriguing meta-finding emerges from the evaluators' scoring patterns. \textit{GPT-5.5} adopts the most conservative evaluation stance (average 3.86/5), while \textit{Qwen 3.6 Max} is the most generous (4.74/5). \textit{Claude Opus 4.7} provides the most balanced and discriminating assessments. This divergence mirrors the behavioral personality profiling in Section 4.2, pointing to that assessment biases are a fundamental property of model architectures.

\textbf{Question Clarity as an Improvement Target.} The \textit{Question Clarity} dimension receives relatively lower scores across all evaluators (average 3.96/5), pointing to that while the scientific content is careful, the phrasing and presentation could benefit from further refinement.

\begin{figure}[t]
 \centering
 \includegraphics[width=1\linewidth]{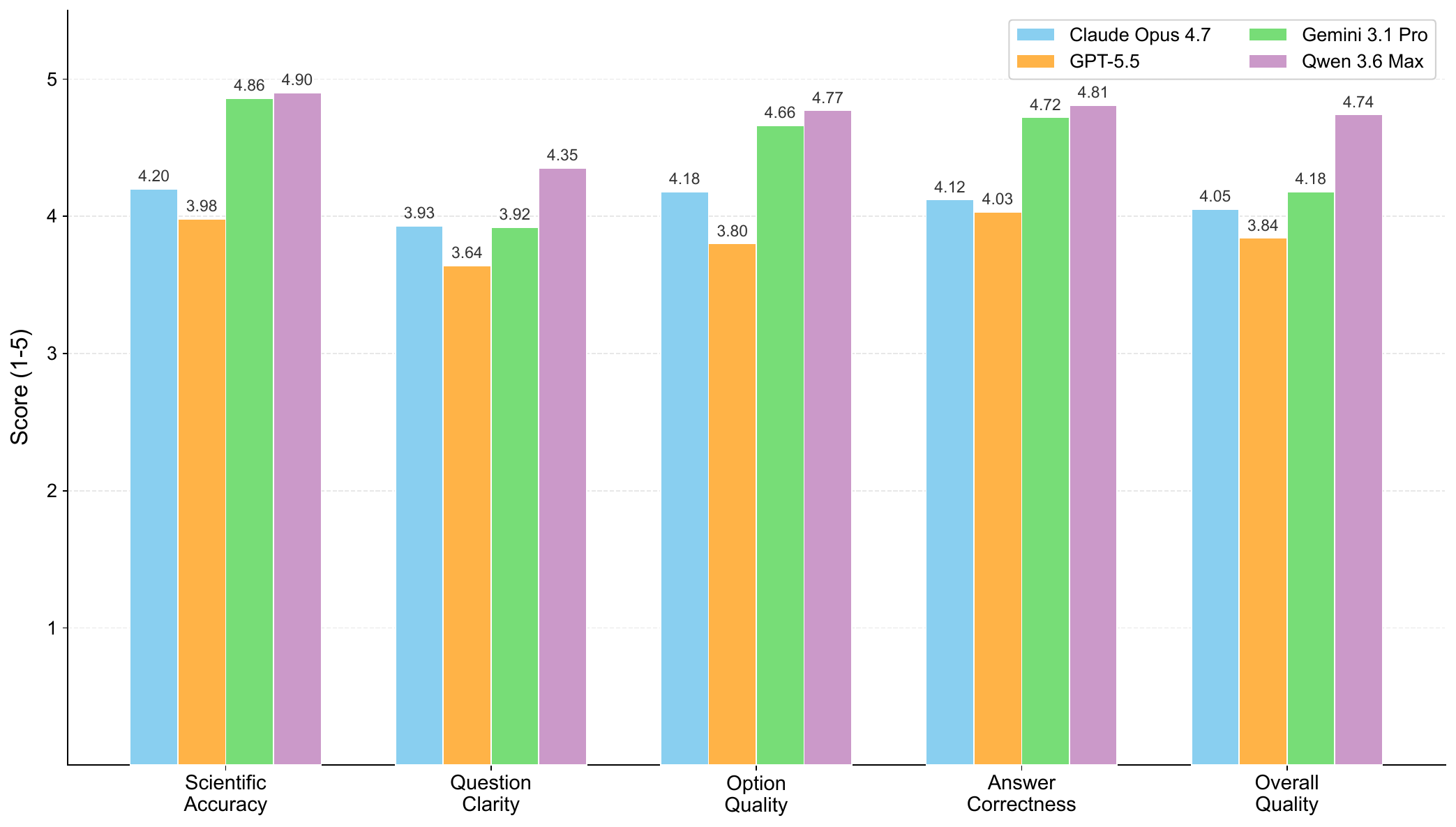}
 \caption{Cross-model quality assessment of SPM-Bench questions.}
 \label{fig:quality_eval}
\end{figure}

\subsection{Field Coverage and Generalizability}

A common concern regarding field-specific benchmarks is whether their findings generalize beyond the narrow field they target. We address this along three dimensions.

\textbf{Broad Sub-field Coverage.} Although framed around microscopy-based science, SPM-Bench spans 538 distinct sub-fields across condensed matter physics, surface science, nanotechnology, materials science, catalysis, superconductivity, energy materials, and more. As shown in Table~\ref{tab:field_coverage}, the top 10 categories cover 69.7\% of questions, while the remaining 528 sub-fields account for 30.3\%---yielding one new sub-field per five questions on average. This field density (538/2,703) exceeds that of general benchmarks such as MMMU (30 fields/2,289 questions) and ScienceQA (50 fields/21,208 questions).

\begin{table}[htbp]
\centering
\caption{Field Coverage of SPM-Bench}
\label{tab:field_coverage}
\small
\resizebox{\columnwidth}{!}{%
\begin{tabular}{lrrr}
 \toprule
 \textbf{Field Category} & \textbf{Questions} & \textbf{Share} & \textbf{Sub-fields} \\
 \midrule
 Condensed Matter Physics & 967 & 35.8\% & 45+ \\
 Surface Science & 596 & 22.0\% & 30+ \\
 Nanotechnology & 121 & 4.5\% & 15+ \\
 Electron Microscopy & 82 & 3.0\% & 10+ \\
 Magnetism \& Spintronics & 34 & 1.3\% & 8+ \\
 Materials Science & 22 & 0.8\% & 12+ \\
 Electrocatalysis & 17 & 0.6\% & 5+ \\
 Heterogeneous Catalysis & 16 & 0.6\% & 4+ \\
 Superconductivity & 15 & 0.6\% & 8+ \\
 Energy Materials & 13 & 0.5\% & 6+ \\
 Other (528 fields) & 820 & 30.3\% & 395+ \\
 \midrule
 \textbf{Total} & \textbf{2,703} & \textbf{100\%} & \textbf{538} \\
 \bottomrule
\end{tabular}%
}
\end{table}

\textbf{Transferable Cognitive Skills.} SPM-Bench evaluates five reasoning capabilities that transfer across all scientific disciplines: (1) \textit{Visual pattern recognition} from complex images (applicable to medical imaging and defect detection); (2) \textit{Multimodal inference} combining visual and textual evidence (scientific paper comprehension); (3) \textit{Hypothesis generation} from observational data (experimental design); (4) \textit{Cross-field knowledge integration} spanning physics, chemistry, and materials science (interdisciplinary research); and (5) \textit{Uncertainty reasoning} under incomplete information (risk assessment). These skills are field-agnostic and directly map to broader scientific reasoning tasks.

\textbf{Extensible Synthesis Pipeline.} The automated 8-step pipeline that produces SPM-Bench is inherently field-agnostic. The Anchor-Gated Sieve filter, VLM-based coordinate inference, local high-fidelity cropping, and adversarial quality control operate on figure-text pairs from any scientific paper. Switching to a new field (e.g., cryo-EM, X-ray crystallography, or astronomical imaging) requires only updating the QA generation prompt templates---the infrastructure remains unchanged. This makes SPM-Bench not just a benchmark but a \textit{blueprint} for constructing deep-dive evaluations in any figure-rich scientific field.

\bibliographystyle{ACM-Reference-Format}
\bibliography{references} 
\clearpage

\appendix
\section{Details in SPM-Bench}
As shown in Figure~\ref{fig:distribution} and detailed in Table~\ref{tab:field_coverage}, SPM-Bench comprises 2,703 high-quality multimodal questions, each paired with a unique, high-resolution microscopy image. 
The dataset maintains a robust balance across four pillar characterization techniques: Transmission Electron Microscopy (TEM, $n=955$), Scanning Tunneling Microscopy (STM, $n=752$), Scanning Electron Microscopy (SEM, $n=558$), and Atomic Force Microscopy (AFM, $n=438$). 
This distribution ensures that the benchmark covers many spatial scales and physical observables, ranging from the atomic-scale electronic density of states (LDOS) in STM to the mesoscopic morphological features and phase transitions in SEM.

\textbf{Taxonomic Hierarchy and Semantic Coverage.}
The sunburst chart in Figure \ref{fig:distribution} shows the multi-layered taxonomic structure of SPM-Bench. 
Beyond the primary imaging modalities, the dataset is systematically organized into specialized physical fields such as \textit{Electronic Structure}, \textit{Magnetic Materials}, \textit{Topological Insulators}, and \textit{Surface Engineering}. 
The outermost ring highlights the "PhD-level" granularity of the benchmark, encompassing high-entropy scientific concepts including the \textit{Rashba Effect}, \textit{Van Hove Singularities}, \textit{Charge Density Waves (CDW)}, and \textit{Skyrmion Lattice} configurations. 
This dense semantic coverage ensures that a model cannot solve SPM-Bench through simple pattern matching; instead, it must perform careful cross-modal reasoning between subtle visual features and fundamental physical laws.

\textbf{Information Density and Complexity.}
The following statistics further highlight the linguistic and reasoning complexity of the benchmark. 
The average prompt length stands at 3,393.4 tokens, pointing to that each question is embedded in a rich, context-heavy environment synthesized from original arXiv and journal publications. 
Also, the average output length of 1,266.0 tokens signifies that the tasks are specifically designed to elicit long-form, analytical reasoning (Chain-of-Thought) rather than brief, superficial answers. 
By integrating peer-reviewed data from over 825 unique scientific papers (407 from arXiv and 418 from high-impact journals), SPM-Bench gives a high-fidelity "academic purity" that effectively minimizes the risk of data contamination common in general-purpose web-crawled datasets.

To benchmark the cognitive maturity of LLMs, we categorize SPM-Bench into five hierarchical levels based on Bloom’s Taxonomy, mapping tasks from basic perception to advanced scientific synthesis:Level 1 - Recall \& Recognition; Level 2 - Comprehension; Level 3 - Application; Level 4 - Analysis; Level 5 - Evaluation \& Synthesis.

\textbf{Level 1 - Recall \& Recognition:} Identification of fundamental imaging modalities or prominent surface features. \textit{E.g., "Is this image acquired via STM or nc-AFM based on the contrast mechanism?"}
 
\textbf{Level 2 - Comprehension:} Interpretation of localized visual signals within a physical context. \textit{E.g., "What does the bright protrusion at the center of the hexagonal lattice represent (e.g., a single adatom or a vacancy)?"}
 
\textbf{Level 3 - Application:} Utilizing physical laws or structural parameters to derive specific properties. \textit{E.g., "Calculate the moiré periodicity and determine the corresponding twist angle between the layers."}
 
\textbf{Level 4 - Analysis:} Correlating heterogeneous data streams, such as multi-panel images and spectroscopic mappings. \textit{E.g., "Map the dI/dV spectroscopic peaks to specific spatial sites to identify the local density of states (LDOS) modulation."}
 
\textbf{Level 5 - Evaluation \& Synthesis:} Critical judgment of physical hypotheses and experimental design. \textit{E.g., "Which atomic model best explains the observed symmetry breaking in the topography, and what subsequent experiment would confirm this?"}

Besides, to carefully probe the ``reasoning boundaries'' of artificial intelligence, \textit{SPM-Bench} categorizes tasks into a specialized cognitive taxonomy, showing the diverse responsibilities of a research scientist as follows:
\textbf{Expert Visual Understanding (EU)} forces the model to perform quantitative and qualitative decoding of complex visual data;
\textbf{Hypothesis Generation (HG)} evaluates inductive reasoning;
\textbf{Experiment Proposal (EP)} tests deductive capability.
 We present the categorizes in the following appendix section C.

\textbf{Model Naming Convention.} The model identifiers used in this paper follow the official API names provided by each vendor. For clarity, we explain the naming convention below:

\begin{itemize}[leftmargin=*, noitemsep]
 \item \textbf{gpt-5.2-1211-global}: GPT-5.2 series, version dated December 11, 2025, global deployment variant.
 \item \textbf{gpt-51-1113-global}: GPT-5.1 series, version dated November 13, 2025, global deployment variant.
 \item \textbf{qwen3.6-max-preview}: Qwen 3.6 series, ``Max'' tier (flagship reasoning capability), preview release.
 \item \textbf{qwen3.5-plus}: Qwen 3.5 series, ``Plus'' tier (mid-range capability).
 \item \textbf{qwen3-vl-plus}: Qwen 3 Vision-Language series, ``Plus'' tier.
 \item \textbf{qwen3-vl-235b-a22b-thinking}: Qwen 3 VL with 235B total parameters (22B active), test-time reasoning scaling enabled (``thinking'' mode).
 \item \textbf{gemini-3-flash-preview}: Gemini 3 series, ``Flash'' tier (lightweight), preview release.
 \item \textbf{gemini-2.5-flash-image-preview}: Gemini 2.5 series, ``Flash'' tier with image-specific optimization, preview release.
 \item \textbf{claude-opus-4.5}: Claude Opus series, version 4.5 (frontier-tier capability).
 \item \textbf{qwen-vl-max}: Qwen Vision-Language series, ``Max'' tier (highest capability).
 \item \textbf{kimi-k2.5}: Kimi series by Moonshot, K2.5 version (flagship reasoning capability).
 \item \textbf{doubao-seed-2-0-pro}: Doubao Seed series by ByteDance, 2.0 Pro version.
\end{itemize}

\clearpage
\section{Sensitivity Analysis of SIP-F1 Metric}
\label{app:sensitivity}

To further justify the robustness of our proposed \textbf{Strict Imperfection Penalty F1 (SIP-F1)} metric, we provide a quantitative sensitivity analysis across various problem complexities ($|S_c|$), model detection outcomes ($\{TP, FP\}$), and hyperparameter configurations ($\Gamma, \lambda$). 

As shown in Table \ref{tab:sensitivity_analysis}, the metric successfully induces a clear bimodal separation between different model "personalities":
\begin{itemize}[leftmargin=*, noitemsep]
 \item \textbf{First Tier (1.00)} Completely correct. Only models with extremely high reasoning capabilities gather in this region.
 \item \textbf{Second Tier ($\sim$0.4, Pink):} Represents the "Conservative" profile, where the model maintains high precision but incomplete recall.
 \item \textbf{Third Tier ($\sim$0.25, Blue):} Represents the "Aggressive" or "Gambling" profile, where correct selections are heavily penalized by the inclusion of hallucinations ($FP$).
\end{itemize}

\textbf{Mathematical Logic of the Bimodal Field} 
As evidenced by the shaded regions in Table \ref{tab:sensitivity_analysis}, the SIP-F1 metric induces a sharp scoring gradient that differentiates "cautious knowledge" from "reckless speculation." Under our selected configuration ($\Gamma=6, \lambda=0.6$), a model that gives a conservative partial answer (e.g., $|S_c|=2, \{TP=1, FP=0\}$) receives a stable score of \textbf{0.400} (Pink Tier). By contrast, an aggressive response that achieves higher recall but includes a single hallucination (e.g., $|S_c|=2, \{TP=2, FP=1\}$) is heavily penalized, dropping the score to \textbf{0.240} (Blue Tier). This inverse relationship between recall and score in the presence of false positives effectively prevents the "Reward for Gambling" common in standard F1-metrics, making sure that the scoring hierarchy remains aligned with scientific integrity.

\textbf{Impact of the Hallucination Penalty ($\Gamma$)} 
The sensitivity of $\Gamma$ across varied problem complexities ($|S_c|=1$ to 3) shows its role as an "epistemic filter." As $\Gamma$ scales from 1 to 10, the scores for any response containing $FP > 0$ show a rapid power-law decay. By setting $\Gamma=6$, we intentionally create a "Strict Penalty Gap" where the cost of a single error (hallucination) outweighs the benefit of completing the recall. This is particularly visible in the $|S_c|=3$ case: selecting two correct answers with zero errors yields 0.480, while selecting all three correct answers but including one error ($FP=1$) results in a notably lower 0.300. This design forces models to prioritize \textit{Precision} over \textit{Recall}—a core requirement for high-stakes laboratory decision-making.

\textbf{Invariance Across Complexity Scales} 
A critical observation from Table \ref{tab:sensitivity_analysis} is the metric's stability across different question densities. Whether a problem requires a single choice ($|S_c|=1$) or a complex multi-panel synthesis ($|S_c|=3$), the Second and Third Tiers remain anchored at approximately 0.4 and 0.25, respectively. This invariance ensures that the "Scientific Personality" diagnosis—categorizing models as \textit{Wise}, \textit{Conservative}, or \textit{Aggressive}—is not skewed by the inherent difficulty of the task. 

\textbf{Anti-Gaming Robustness} 
The bottom-most rows of each complexity block highlight the metric's resistance to "gaming the benchmark." For models that adopt a "select-all" strategy (represented by high $FP$ counts), the scores converge toward the baseline as $\Gamma$ increases. At our chosen $\Gamma=6$, such behavior is rendered statistically unviable, yielding scores lower than a purely random guess. 

\textbf{Complementarity with Calibration Metrics.} We further examine the relationship between SIP-F1 and Expected Calibration Error (ECE) to show that they capture orthogonal dimensions of model behavior. As shown in Table \ref{tab:sipf1-ece-comparison}, models with a "Wise" personality (\textit{qwen3.5-plus}, \textit{gpt-5.2-1211-global}) achieve both high SIP-F1 ($\geq$0.81) and low ECE ($\leq$0.095), pointing to well-calibrated confidence. By contrast, "Aggressive" models (\textit{gemini-3-flash-preview}) show moderate SIP-F1 (0.710) but high ECE (0.183), showing a mismatch between self-confidence and actual performance. This confirms that SIP-F1 evaluates answer precision while ECE evaluates confidence-accuracy alignment---two complementary axes for diagnosing model reliability.

\textbf{Positive Predictive Bias in "Aggressive" Models.} 
Mid-tier models, exemplified by \textit{gemini-3-flash-preview}, show what we term a \textbf{Positive Predictive Bias}. These models tend to over-select options in multi-select scientific tasks to maximize recall, effectively "gaming" traditional F1 metrics. However, through the lens of SIP-F1, this behavior is exposed as a lack of logical grounding. Statistically, this bias shows as a misalignment between high self-reported confidence and low precision. In high-stakes physical discovery, such aggressive predicting is hazardous, as it produces false-positive "discoveries" by speculating on ambiguous visual signals.

\textbf{Epistemic Uncertainty Awareness in "Wise" Models.} 
By contrast, frontier models like the \textit{GPT-5} series show \textbf{High Epistemic Uncertainty Awareness}. These models are not merely "smarter" but are better calibrated; they show a advanced ability to quantify the limits of their own knowledge. When faced with high-entropy physical scenarios (e.g., symmetry-breaking in disordered lattices), they adopt a \textbf{Conservative Strategy}, prioritizing precision over speculative recall. This behavior reflects a deep logical internal consistency, where the model's confidence is strongly correlated with the task's intrinsic difficulty. Such models act as reliable scientific collaborators because they "know when they do not know," thereby minimizing the risk of cascading errors in autonomous research pipelines.

\begin{table}[htbp]
\centering
\caption{Quantitative sensitivity analysis of the SIP-F1 metric across varied problem complexities ($|S_c|$), detection outcomes ($\{TP, FP\}$), and parameter configurations ($\Gamma, \lambda$). Highlighted cells distinguish performance tiers: blue points to scores approximately at 0.25 (Third Tier), while pink points to scores approximately at 0.4 (Second Tier).}
\label{tab:sensitivity_analysis}
\small 
\renewcommand{\arraystretch}{1.2} 
\setlength{\tabcolsep}{0pt} 

\begin{tabular*}{\columnwidth}{@{\extracolsep{\fill}} l l c cccccc @{}}
\toprule
$|S_c|$ & $\{TP, FP\}$ & $\Gamma$ & $\lambda=0.3$ & 0.4 & 0.5 & 0.6 & 0.7 & 0.8 \\ 
\midrule

% Sc = 1
\multirow{6}{*}{\textbf{1}} & \multirow{6}{*}{$\{1, 1\}$} 
 & 1 & 0.200 & \cellcolor{bluebg}\textbf{0.267} & 0.333 & \cellcolor{pinkbg}\textbf{0.400} & 0.467 & 0.533 \\
& & 2 & 0.150 & 0.200 & \cellcolor{bluebg}\textbf{0.250} & 0.300 & 0.350 & \cellcolor{pinkbg}\textbf{0.400} \\
& & 4 & 0.100 & 0.133 & 0.167 & 0.200 & \cellcolor{bluebg}\textbf{0.233} & \cellcolor{bluebg}\textbf{0.267} \\
& & 6 & 0.075 & 0.100 & 0.125 & 0.150 & 0.175 & 0.200 \\
& & 8 & 0.060 & 0.080 & 0.100 & 0.120 & 0.140 & 0.160 \\
& & 10 & 0.050 & 0.067 & 0.083 & 0.100 & 0.117 & 0.133 \\ 
\midrule

% Sc = 2
\multirow{7}{*}{\textbf{2}} & $\{1, 0\}$ & -- & 0.200 & \cellcolor{bluebg}\textbf{0.267} & 0.333 & \cellcolor{pinkbg}\textbf{0.400} & 0.467 & 0.533 \\ 
\cmidrule(lr){2-9}
& \multirow{6}{*}{$\{2, 1\}$} 
 & 1 & \cellcolor{bluebg}\textbf{0.240} & 0.320 & \cellcolor{pinkbg}\textbf{0.400} & 0.480 & 0.560 & 0.640 \\
& & 2 & 0.200 & \cellcolor{bluebg}\textbf{0.267} & 0.333 & \cellcolor{pinkbg}\textbf{0.400} & 0.467 & 0.533 \\
& & 4 & 0.150 & 0.200 & \cellcolor{bluebg}\textbf{0.250} & 0.300 & 0.350 & \cellcolor{pinkbg}\textbf{0.400} \\
& & 6 & 0.120 & 0.160 & 0.200 & \cellcolor{bluebg}\textbf{0.240} & 0.280 & 0.320 \\
& & 8 & 0.100 & 0.133 & 0.167 & 0.200 & \cellcolor{bluebg}\textbf{0.233} & \cellcolor{bluebg}\textbf{0.267} \\
& & 10 & 0.086 & 0.114 & 0.143 & 0.171 & 0.200 & 0.229 \\ 
\midrule

% Sc = 3
\multirow{8}{*}{\textbf{3}} & $\{2, 0\}$ & -- & \cellcolor{bluebg}\textbf{0.240} & 0.320 & \cellcolor{pinkbg}\textbf{0.400} & 0.480 & 0.560 & 0.640 \\ 
& $\{1, 0\}$ & -- & 0.150 & 0.200 & \cellcolor{bluebg}\textbf{0.250} & 0.300 & 0.350 & \cellcolor{pinkbg}\textbf{0.400} \\ 
\cmidrule(lr){2-9}
& \multirow{6}{*}{$\{3, 1\}$} 
 & 1 & \cellcolor{bluebg}\textbf{0.257} & 0.343 & \cellcolor{pinkbg}\textbf{0.429} & 0.514 & 0.600 & 0.686 \\
& & 2 & 0.214 & 0.286 & 0.357 & \cellcolor{pinkbg}\textbf{0.429} & 0.500 & 0.571 \\
& & 4 & 0.180 & \cellcolor{bluebg}\textbf{0.240} & 0.300 & 0.360 & \cellcolor{pinkbg}\textbf{0.420} & 0.480 \\
& & 6 & 0.150 & 0.200 & \cellcolor{bluebg}\textbf{0.250} & 0.300 & 0.350 & \cellcolor{pinkbg}\textbf{0.400} \\
& & 8 & 0.130 & 0.173 & 0.217 & \cellcolor{bluebg}\textbf{0.261} & 0.304 & 0.348 \\
& & 10 & 0.113 & 0.150 & 0.188 & 0.225 & \cellcolor{bluebg}\textbf{0.263} & 0.300 \\ 
\bottomrule
\end{tabular*}
\end{table}

\section{Model Performance Across Different Fields}
\textbf{Deduction Beyond Perception} An intriguing observation arises from the performance distribution of the frontier \textit{gpt-5.2-1211-global}. Its peak performance is achieved in \textit{Energy Materials} (0.7600) and \textit{Surface Chemical Reactions} (0.7429)---fields that demand careful causal reasoning and the synthesis of multi-panel data (e.g., correlating STM topography with local spectroscopy). By contrast, its score in \textit{Morphology Analysis} (0.7156) is relatively lower, despite this being a more inherently visual'' task. This points to that the scaling of large multimodal models primarily augments their \textbf{scientific deductive reasoning} rather than their raw visual acuity. The model is no longer merely recognizing'' a cluster; it is ``calculating'' its structural and chemical implications through an internal physics-aligned logic.

\begin{figure}
 \centering
 \includegraphics[width=0.8\columnwidth]{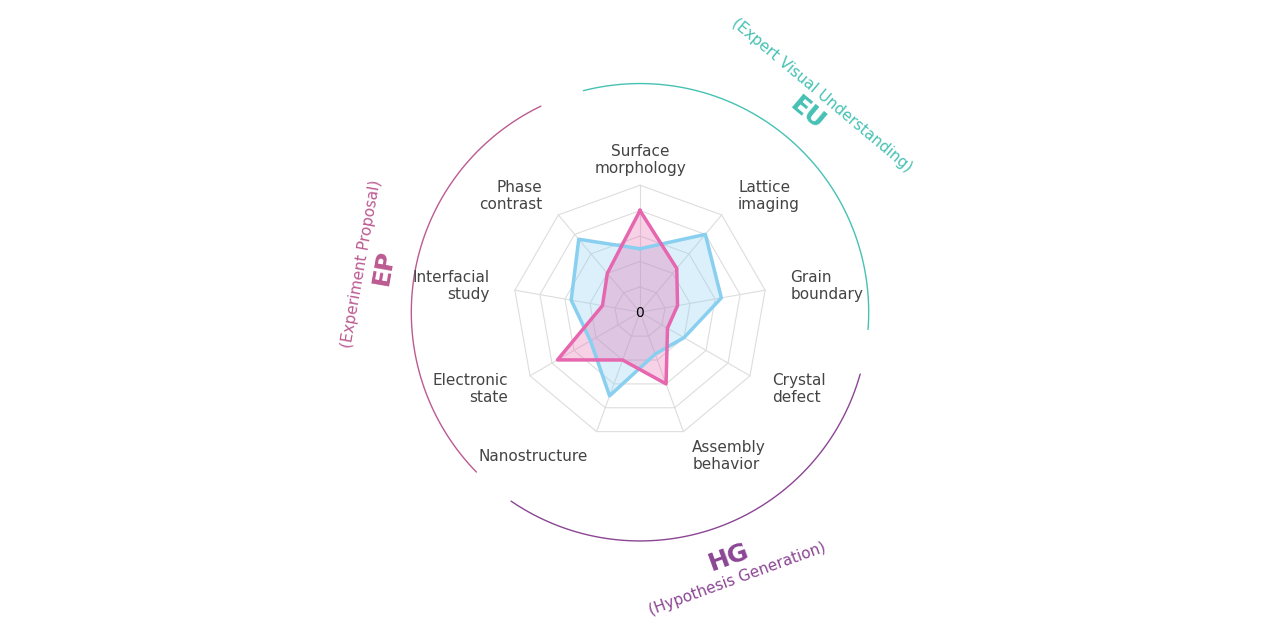}
 \caption{Field-specific competence envelopes across model families. Unlike Figure~\ref{fig:distribution} which organizes questions by microscopy type (TEM/STM/SEM/AFM), this chart categorizes by scientific field because field reflects distinct cognitive reasoning challenges (e.g., visual pattern recognition vs. quantum-mechanical inference), whereas microscopy type is a technical dimension that cuts across cognitive tasks.}
 \label{fig:radar}
\end{figure}

\textbf{Stability in Frontier Competence} As visualized in Figure~\ref{fig:radar}, the \textit{gpt-5} series shows an almost perfect circular'' profile at high radii. This is a state of \textbf{balanced expert capability}, where the model performs consistently across disparate fields, from \textit{Magnetic Materials} (0.7013) to \textit{Electronic Structure} (0.6929). By contrast, mid-tier models like \textit{gemini-2.5-flash-image-preview} and \textit{qwen3-vl-plus} show notably spiky'' or restricted envelopes. These models frequently collapse in high-abstraction fields like \textit{Defect Engineering}, where identifying an atomic vacancy or a dopant site (scoring as low as 0.2917 for \textit{gemini-2.5-flash-image-preview}) requires the model to deduce symmetry-breaking rules from subtle contrast variations---a task that remains a definitive bottleneck for VLMs lacking sufficient reasoning-optimization.

\textbf{Bottlenecks in Quantum Fields} Despite the general progress, \textit{Topological States and Quantum Materials} remains one of the most formidable fields for all evaluated models. While the mid-tier \textit{qwen3-vl-plus} hovers at a baseline of 0.4712, the frontier \textit{gpt-5.2-1211-global} manages to reach 0.7043. The ability of the top-tier series to maintain scores above 0.70 in this area confirms that advanced internal verification and latent physical knowledge are essential for decoding the high-dimensional data inherent in quantum phenomena. However, the persistent gap between \textit{Morphology} and \textit{Defect/Topology} across the entire spectrum highlights that the transition from visual characterization'' to quantum state inference'' is the primary frontier for the next generation of AI4Science models.
The radar charts illustrate a comparative performance analysis of three major Large Language Model (LLM) families—GPT, Gemini, and Qwen—across eight specialized physical science fields. The results show distinct patterns in generational scaling, reasoning capabilities, and field-specific sensitivities.

\textbf{Impact of Reasoning Mechanisms on Scientific Accuracy}
The comparison between qwen3-vl-235b-thinking and qwen3-vl-plus (Bottom) highlights the deep impact of specialized reasoning architectures. The "Thinking" version shows a significant "area expansion" in the radar plot, particularly in the Electronic, Morphology, and Crystal fields. This points to that the "thinking" (chain-of-thought) mechanism allows the model to better parse the spatial hierarchy of scientific figures, leading to a more accurate mapping of visual evidence to physical laws. Conversely, the Defect and Energy fields remain common bottlenecks, likely due to the inherent complexity of identifying localized irregularities and low-contrast features in microscopy data.

\textbf{Generational Leap and Perceptual Evolution}
The Gemini series (Middle) shows the most dramatic generational performance shift. While \textit{gemini-2.5-flash-image-preview} shows a constrained profile centered around 0.2–0.5, \textit{gemini-3-flash-preview} shows a robust expansion across all axes, nearly doubling the performance in the Electronic and Topology sectors. This leap points to a fundamental upgrade in the model's visual encoder, enabling it to transition from basic pattern recognition to a more advanced "semantic understanding" of scientific layouts. However, even with this progress, a noticeable "dip" in the Defect axis persists, pointing to that characterizing microscopic imperfections remains a cross-model challenge in the current state of VLMs.

\clearpage

% --- Example 1: SEM ---
\begin{figure*}[t]
\section{Typical QA in SPM-Bench}
\begin{custombox}{SoftPurple}{Example 1: SEM}
\textbf{Question: } The SEM images in panels (a) through (f) depict the structural evolution of a condensed phase system within a liquid medium. Panel (a) shows the initial dominance of \textbf{Phase $\alpha$}, characterized by a curvilinear morphology. The high-magnification inset in panel (e) shows the surface topography of \textbf{Phase $\alpha$} entity. As the system evolves toward panel (f), \textbf{Phase $\beta$}, exhibiting a distinct faceted habit, becomes the sole constituent. Based on the visual evidence provided and the fundamental principles of phase transition kinetics, which of the following statements regarding the observed process are consistent with the visual evidence?

\begin{itemize}[leftmargin=*, itemsep=2pt]
 \item[A.] The transition from Phase $\alpha$ to Phase $\beta$ is a solution-mediated process driven by a chemical potential gradient, where the higher solubility of the metastable Phase $\alpha$ helps the nucleated growth of the thermodynamically stable Phase $\beta$.
 \item[B.] The granular substructure observed in the inset of (e) identifies Phase $\alpha$ as a mesocrystalline assembly of nanocrystalline subunits, pointing to a non-classical crystallization pathway involving the oriented attachment of primary particles.
 \item[C.] The morphological evolution is a topotactic solid-state transition, where the faceted geometry in (f) emerges through the collective migration of coherent grain boundaries within the precursor Phase $\alpha$ volume.
 \item[D.] The coexistence of Phase $\alpha$ and Phase $\beta$ in panels (b) through (e) is indicative of a first-order phase transition where the global transformation rate is governed by the competitive rates of dissolution and secondary nucleation.
 \item[E.] The transition is a purely curvature-driven Ostwald ripening process within a single-phase system, where the faceted habit in (f) represents the minimization of the surface-to-volume ratio of the initial objects in (a).
 \item[F.] The inset in (e) shows that Phase $\alpha$ is an amorphous precursor, and the transition to (f) occurs via a glass-to-crystal transition that bypasses the liquid-phase chemical equilibrium.
\end{itemize}
\medskip 
\textbf{Answer: } \fbox{A, B, D}
\end{custombox}

% --- Metadata Box for Example 1 ---
\vspace{-0.7mm} 
\begin{tcolorbox}[
 colback=SoftPurple!3!white, 
 colframe=SoftPurple,  
 arc=3mm,   
 boxrule=1.5pt,  
 left=12pt, right=12pt, 
 top=8pt, bottom=8pt,  
]
 \begin{minipage}[c]{0.4\linewidth}
 \centering
 \includegraphics[width=0.7\linewidth]{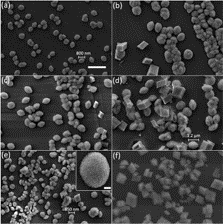}
 \end{minipage}
 \hfill
 \begin{minipage}[c]{0.52\linewidth}
 \normalsize \raggedright 
 \setlength{\parskip}{4.5pt} 

 {\bfseries Level:} PhD \par
 {\bfseries Discipline:} Condensed Matter Physics \par
 {\bfseries Field:} Thermodynamics and Phase Transitions \par
 {\bfseries Question Type:} HG \\
 \end{minipage}
\end{tcolorbox}

\vspace{1cm}
\end{figure*}

% --- Example 2: TEM ---
\begin{figure*}[t]
\begin{custombox}{SoftBlue}{Example 2: TEM}
\textbf{Question: }A multicomponent inorganic specimen is characterized using various electron-beam modalities in panels (a-k). Panel (f) is an incoherent imaging mode, while panels (c-e) use a coherent phase-contrast mode. Panels (g-k) display the spatial distribution of Species 1-5 (labeled G-K). Based on the visual evidence and the underlying physics of electron-matter interactions, which of the following statements regarding the observed process and data interpretation are consistent with the evidence?

\begin{itemize}[leftmargin=*, itemsep=2pt]
 \item[A.] The spatial distribution of Species H and I in panels (h) and (i) points to preferential segregation at the grain boundaries visible in (b), pointing to a solute-drag mechanism during crystallization.
 \item[B.] The intensity variations in the micrograph (f) are purely a function of the atomic number ($Z^n$) of the constituent species, as the imaging mode effectively filters out all thickness-dependent dynamical diffraction effects.
 \item[C.] The contrast in panels (c--e) is governed by the interference of the transmitted and diffracted beams, where the image intensity is a convolution of the specimen's projected potential and the oscillatory Contrast Transfer Function (CTF) of the objective lens.
 \item[D.] The measured $d$-spacings in (d) and (e) (0.174~nm and 0.195~nm) provide sufficient evidence to confirm a transition from a low-symmetry monoclinic phase to a high-symmetry $Fd\bar{3}m$ cubic phase across the entire specimen.
 \item[E.] The lower signal-to-noise ratio observed for Species K in panel (k) relative to Species G in (g) is primarily due to the notably higher X-ray fluorescence yield of light elements compared to transition metals.
 \item[F.] Deviations in the lattice periodicities observed in (d) and (e) from those of a stoichiometric reference can be used to quantify the local concentration of Species H and I via Vegard’s Law, provided the system maintains a homogeneous solid solution.
\end{itemize}
\medskip 
\textbf{Answer: } \fbox{C, F}
\end{custombox}

% --- Metadata Box for Example 2 ---
\vspace{-0.7mm} 
\begin{tcolorbox}[
 colback=SoftBlue!3!white, 
 colframe=SoftBlue,  
 arc=3mm,   
 boxrule=1.5pt,  
 left=12pt, right=12pt, 
 top=8pt, bottom=8pt,  
]
 \begin{minipage}[c]{0.4\linewidth}
 \centering
 \includegraphics[width=1\linewidth]{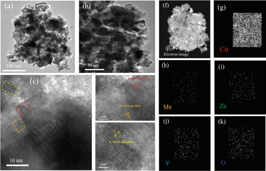} 
 \end{minipage}
 \hfill
 \begin{minipage}[c]{0.52\linewidth}
 \normalsize \raggedright 
 \setlength{\parskip}{4.5pt} 

 {\bfseries Level:} PhD \par
 {\bfseries Discipline:} Surface Science \par
 {\bfseries Field:} Nanomaterials Characterization \par
 {\bfseries Question Type:} EP \\
 \end{minipage}
\end{tcolorbox}
\vspace{1cm}
\end{figure*}
% --- Example 3: STM ---
\begin{figure*}[t]
\begin{custombox}{SoftPink}{Example 3: STM}
\textbf{Question: }The provided scanning tunneling microscopy (STM) images (a--f) characterize the structural and electronic evolution of a 2D overlayer (labeled $\Lambda$) supported on a metallic substrate ($\Sigma_1$) following the introduction of an atomic species ($\chi$). Panel (c) depicts the ``as-grown'' state of $\Lambda$ on $\Sigma_1$. Panels (d), (e), and (f) illustrate various stages of the incorporation of $\chi$ into the interface. Specifically, panel (e) captures a single field of $\Lambda$ spanning two distinct interfacial environments. Based on the visual evidence and the fundamental principles of surface physics, which of the following statements regarding the observed process are consistent with the data?

\begin{itemize}[leftmargin=*, itemsep=2pt]
 \item[A.] The apparent height increase observed in the $\Lambda/\chi$ regions of image (e) is a purely topographic measurement of the $\chi$ monolayer thickness, independent of the local tunneling decay constant $\kappa$.
 \item[B.] The emergence of the short-period periodic modulation in (f) relative to (c) implies that the lattice mismatch between $\Lambda$ and the underlying layer increases notably upon the introduction of $\chi$, as the superlattice periodicity $\lambda$ is inversely proportional to the mismatch $\delta$.
 \item[C.] The transition from state (c) to (f) signifies an electronic decoupling of $\Lambda$ from the substrate, shifting the system from a strongly hybridized chemisorbed state to a physisorbed state where the intrinsic electronic structure of the overlayer is largely restored.
 \item[D.] The contrast in (c) is darker than in (f) primarily because the work function ($\phi$) of the $\Lambda/\Sigma_1$ interface is notably higher than that of the $\Lambda/\chi/\Sigma_1$ interface, thereby improving the tunneling probability in the latter.
 \item[E.] The spatial configuration in panel (e) shows that the species $\chi$ undergoes a surfactant-mediated growth mode, where $\chi$ atoms reside exclusively on the exterior surface of $\Lambda$ to minimize interface energy.
 \item[F.] The visual evidence in (e) and the schematic in (d) confirm that the intercalation of $\chi$ is an edge-mediated process, where species $\chi$ infiltrates the interface at the boundaries of the $\Lambda$ fields.
\end{itemize}
\medskip 
\textbf{Answer: } \fbox{B, C, F}
\end{custombox}

% --- Metadata Box for Example 3 ---
\vspace{-0.7mm} 
\begin{tcolorbox}[
 colback=SoftPink!3!white, 
 colframe=SoftPink,  
 arc=3mm,   
 boxrule=1.5pt,  
 left=12pt, right=12pt, 
 top=8pt, bottom=8pt,  
]
 \begin{minipage}[c]{0.4\linewidth}
 \centering
 \includegraphics[width=\linewidth]{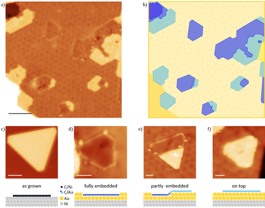} 
 \end{minipage}
 \hfill
 \begin{minipage}[c]{0.52\linewidth}
 \normalsize \raggedright 
 \setlength{\parskip}{4.5pt} 

 {\bfseries Level:} PhD \par
 {\bfseries Discipline:} Condensed Matter Physics \par
 {\bfseries Field:} Graphene Nanoislands \par
 {\bfseries Question Type:} EU \\
 \end{minipage}
\end{tcolorbox}
\vspace{1cm}
\end{figure*}

% --- Example 4: AFM ---
\begin{figure*}[t]
\begin{custombox}{SoftOrange}{Example 4: AFM}
\textbf{Question: }The provided figure illustrates the formation of a surface coordination layer on a Cu(110) substrate, characterized by STM (a, b) and nc-AFM (c). Panel (c) displays a frequency shift ($\Delta f$) map acquired with a CO-functionalized tip, showing sub-molecular features within a $c(6 \times 2)$ unit cell. The experimental distance between the prominent bright protrusions is measured at $3.73~\text{\AA}$, which closely matches the $3.66~\text{\AA}$ distance between oxygen atoms in the carboxylate ligands shown in the structural model (e). Considering the contrast mechanism in non-contact AFM operating in the repulsive regime with a flexible probe, which of the following statements correctly describes the physical origin of the observed features and the implications for the surface electronic structure?

\begin{itemize}[leftmargin=*, itemsep=2pt]
 \item[A.] The bright features in (c) represent the local density of states (LDOS) at the Fermi level, dominated by the metal $d$-orbitals of the coordination center.
 \item[B.] The contrast is primarily governed by Pauli repulsion between the CO tip's frontier orbitals and the electron density of the carboxylate oxygen atoms.
 \item[C.] The $3.73~\text{\AA}$ distance is a direct measurement of the Cu--Cu distance in the coordination layer, pointing to a metallic bond.
 \item[D.] The frequency shift $\Delta f$ is independent of the tip-sample distance in the repulsive regime, allowing for height-independent imaging.
 \item[E.] The sub-molecular resolution is a result of the Kondo effect at the metal-organic interface, which improves the tunneling probability.
 \item[F.] The observed contrast arises from the electrostatic attraction between the CO dipole and the partial positive charge on the hydrogen atoms.
\end{itemize}
\medskip 
\textbf{Answer: } \fbox{B}
\end{custombox}

% --- Metadata Box for Example 4 ---
\vspace{-0.7mm} 
\begin{tcolorbox}[
 colback=SoftOrange!3!white, 
 colframe=SoftOrange,  
 arc=3mm,   
 boxrule=1.5pt,  
 left=12pt, right=12pt, 
 top=8pt, bottom=8pt,  
]
 \begin{minipage}[c]{0.4\linewidth}
 \centering
 \includegraphics[width=\linewidth]{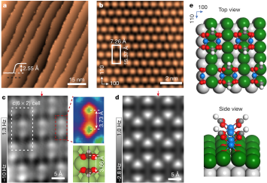} 
 \end{minipage}
 \hfill
 \begin{minipage}[c]{0.52\linewidth}
 \normalsize \raggedright 
 \setlength{\parskip}{4.5pt} 

 {\bfseries Level:} PhD \par
 {\bfseries Discipline:} Surface Science \par
 {\bfseries Field:} Surface Reconstruction \par
 {\bfseries Question Type:} HG \\
 \end{minipage}
\end{tcolorbox}
\end{figure*}

\section{Licensing and Data Usage Terms}
\label{app:licensing}

SPM-Bench is released under a dual-licensing framework to ensure both open academic access and protection of third-party copyright.

\textbf{Source Code License.} All source code in the SPM-Bench repository (including data loaders, evaluation metrics, and synthesis pipeline) is released under the \textbf{Apache License 2.0}. This permits unrestricted use, modification, and distribution for both academic and commercial purposes, subject to the terms of the Apache 2.0 license.

\textbf{Dataset License.} The SPM-Bench dataset (including questions, answers, and metadata) is licensed under the \textbf{Creative Commons Attribution-NonCommercial-ShareAlike 4.0 International License} (CC BY-NC-SA 4.0). This permits academic sharing and adaptation under the conditions of attribution, non-commercial use, and share-alike.

\textbf{Image Copyright.} The 2,703 images in SPM-Bench are extracted from 825 academic papers published between 2023--2025, including 407 arXiv preprints and 418 journal articles. The original copyright of individual images remains with their respective publishers (APS, ACS, Elsevier, Springer Nature, Wiley, RSC, IOP Publishing, etc.). These images are included under \textbf{fair use} principles for academic research:

\begin{itemize}[leftmargin=*, noitemsep]
 \item \textbf{Transformative Use:} Images are repurposed from scientific communication into an AI evaluation benchmark, testing model reasoning capabilities rather than communicating scientific findings.
 \item \textbf{Non-Commercial:} The CC BY-NC-SA 4.0 license explicitly prohibits commercial use.
 \item \textbf{Attribution:} A machine-readable image provenance file (\texttt{data/image\_sources.json}) traces each image back to its source paper, including the paper identifier (arXiv ID or DOI), source URL, and copyright holder.
 \item \textbf{No Substitution:} SPM-Bench does not substitute for the original papers. Users must cite the original source papers when using specific images in publications.
\end{itemize}

\textbf{Permitted Use Cases:}
\begin{itemize}[leftmargin=*, noitemsep]
 \item Academic research (benchmarking LLM/VLM performance, publishing comparative studies)
 \item Education (classroom instruction, workshop demonstrations)
 \item Non-commercial AI safety research (evaluating model hallucination, calibration, and reasoning boundaries)
\end{itemize}

\textbf{Prohibited Use Cases:}
\begin{itemize}[leftmargin=*, noitemsep]
 \item Commercial use (product evaluation, commercial benchmark reports)
 \item Model training (using SPM-Bench content for training or fine-tuning AI models)
 \item Standalone image redistribution (extracting and redistributing individual images outside the benchmark context)
\end{itemize}

\textbf{Citation Requirement.} Any publication using SPM-Bench must cite: Xiao et al., ``SPM-Bench: Benchmarking Large Language Models for Scanning Probe Microscopy,'' arXiv:2602.22971, 2026. When using specific images, users must additionally cite the original source paper as listed in \texttt{data/image\_sources.json}.

Full terms of use are available at \texttt{TERMS\_OF\_USE.md} in the SPM-Bench repository.

\end{document}